\documentclass[journal,comsoc]{IEEEtran}
\usepackage[T1]{fontenc}
\ifCLASSINFOpdf
\else
\fi
\usepackage[cmintegrals]{newtxmath}
\usepackage{amsmath}
\usepackage[ruled]{algorithm2e}
\usepackage{times}
\usepackage{epsfig}
\usepackage{graphicx}
\usepackage{amsmath}

\usepackage{amssymb}
\usepackage{multirow}
\usepackage{verbatim}
\usepackage{bm}
\usepackage{framed}
\usepackage{makecell}
\usepackage{booktabs}
\usepackage{multirow}
\usepackage{url}

\begin{document}	
	
\title{Towards Transferable Adversarial Attack against Deep Face Recognition}
\author{Yaoyao Zhong,
	Weihong Deng
	\thanks{The authors are with the Pattern Recognition and Intelligent System
		Laboratory, School of Artificial Intelligence, Beijing University of Posts and Telecommunications, Beijing 100876, China (e-mail: zhongyaoyao@bupt.edu.cn; whdeng@bupt.edu.cn). Weihong Deng is the corresponding author.}
}
\maketitle

\begin{abstract}
Face recognition has achieved great success in the last five years due to the development of deep learning methods. However, deep convolutional neural networks (DCNNs) have been found to be vulnerable to adversarial examples. In particular, the existence of transferable adversarial examples can severely hinder the robustness of DCNNs since this type of attacks can be applied in a fully black-box manner without queries on the target system. In this work, we first investigate the characteristics of transferable adversarial attacks in face recognition by showing the superiority of feature-level methods over label-level methods. Then, to further improve transferability of feature-level adversarial examples, we propose DFANet, a dropout-based method used in convolutional layers, which can increase the diversity of surrogate models and obtain ensemble-like effects. Extensive experiments on state-of-the-art face models with various training databases, loss functions and network architectures show that the proposed method can significantly enhance the transferability of existing attack methods. Finally, by applying DFANet to the LFW database, we generate a new set of adversarial face pairs that can successfully attack four commercial APIs without any queries. This TALFW database is available to facilitate research on the robustness and defense of deep face recognition.
\end{abstract}
\IEEEpeerreviewmaketitle


\section{Introduction}

\IEEEPARstart{D}{eep} convolutional neural networks (DCNNs) have achieved great success in face recognition~\cite{sun2014deep,Schroff2015FaceNet,wen2016discriminative,liu2017sphereface,chen2017noisy,Wang2018CosFace,deng2019arcface}. In unconstrained environments, the face recognition performance is reaching saturation levels on several benchmarks, \emph{e.g.}, LFW~\cite{LFWTech}, MegaFace~\cite{kemelmacher2016megaface} and IJB-A~\cite{klare2015pushing}. Even in the real security certificate environment, which requires a very low false positive rate~\cite{maze2018iarpa}, state-of-art performance has been achieved~\cite{deng2019arcface}. Although deep face models have already surpassed human performance~\cite{zhou2015naive,deng2017fine} on some benchmarks, we must keep in mind that benchmarks may not be able to capture realistic performance~\cite{goodfellow2018defense}. Because of the remarkable performance, face recognition applications can be numerous as well as diverse in our daily lives. Face recognition can be widely used in security agencies, law enforcement agencies, the airline industry, the banking and securities industries and so on, which further increases the demand for security in deep face models. 

\begin{figure}[htbp]
	\center
	\includegraphics[width=1\linewidth]{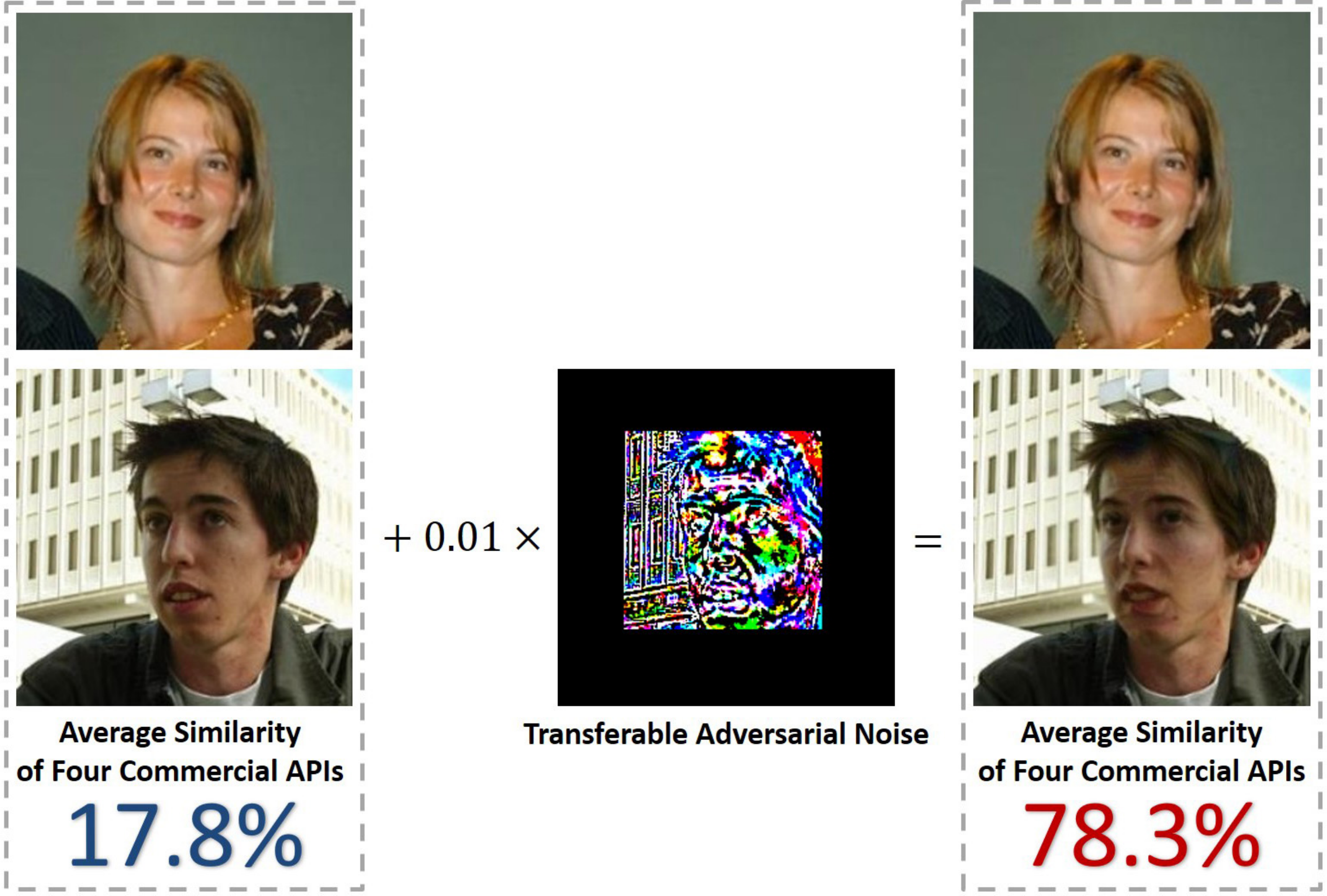}
	\caption{A transferable adversarial face pair. The original face pair is shown on the left, which contains two face images of different identities. Although perturbation of the modified face image is imperceptible, the similarity score of four commercial APIs changed significantly.}
	\label{fig:intro_pair}
\end{figure}

Researchers found that the basis of deep face models, DCNNs, can be easily fooled by adversarial examples, which are modified from original test images by adding noise not perceptible to humans~\cite{Szegedy14,goodfellow2014explaining}. Moreover, adversarial examples are transferable in different models~\cite{liu2016delving,papernot2017practical,dong2018boosting,xie2019improving}, which means that black-box attacks can be launched from the local surrogate models without having information on the remote target systems, including network architectures, model parameters, training databases and defensive methods. 

Face recognition systems are not robust as we think when test images are not drawn from the same distribution as the training images. Previous works have already demonstrated the existence of the vulnerability of deep face models~\cite{sharif2016accessorize,song2018attacks,dong2019efficient}. However, previous works in white-box settings~\cite{sharif2016accessorize,rozsa2017lots} assume that the attacker has full access to the target models. Query-based methods in black-box settings~\cite{dong2019efficient} require a large number of queries, which could be detected relatively easily by the target system. Both settings are not practical in real-world situations because the remote target system can neither release the network information proactively nor allow a large number of queries on a face pair. 

In contrast, transferable adversarial attacks can severely hinder robustness since this type of attack can be applied in a fully black-box manner without any queries on the target system. We demonstrate that, with this type of adversarial attack, state-of-art face models and even commercial APIs can be attacked, as shown in Fig.~\ref{fig:intro_pair}. Both the original face pair and the modified face pair are shown. The face images of the original pair belong to different identities while the modified pair is judged as the same identity by four commercial APIs: Amazon~\cite{Amazon}, Microsoft~\cite{azure}, Baidu~\cite{Baidu} and Face++~\cite{Face++}. Because of the great potential harms of transferable adversarial attacks in deep face recognition, it is valuable to study this issue in a black-box setting without any queries. 

Although recent works have investigated the transferability of adversarial examples~\cite{liu2016delving,papernot2017practical,dong2018boosting,xie2019improving} in DCNNs used for close-set object classification tasks~\cite{deng2009imagenet}, some works remained to be done on transferable adversarial attacks in deep face recognition because of the particularity of deep face models. For a target black-box deep face model, there are a wide range of options for its training databases~\cite{Yi2014CASIA,Guo16MS,cao2018vggface2,wang2018devil}, training loss functions~\cite{sun2014deep,Schroff2015FaceNet,wen2016discriminative,liu2017sphereface,chen2017noisy,Wang2018CosFace,deng2019arcface} and network architectures~\cite{he2016deep,howard2017mobilenets,szegedy2017inception,hu2018squeeze}. Because it has been widely accepted both in academia and industry that these three key pieces of information can improve recognition performance, and there are still many works insisting on designing new and effective ones. The variety of deep face models will undoubtedly increase the difficulty of generating transferable adversarial attacks if we have no knowledge of the target systems.  

The objective of this work is to study transferable adversarial attacks against deep face models, \emph{i.e}., to generate adversarial face pairs from surrogate models and lead the black-box face models to misjudge these generated face pairs by depending only on the transferability without any queries of the target models. Furthermore, to simulate the potential scenarios, we increase the difference between source and target models by attacking deep face models with different training databases, training loss functions, and network architectures. To this end, considering the aforementioned particularity, we first start by investigating the applicable strategies of transferable adversarial attacks against deep face recognition to select a suitable baseline method. To further improve the transferability of adversarial examples, we propose the dropout face attacking network (DFANet), a dropout-based method used in convolutional layers. DFANet can increase the diversity of surrogate models and prevent overfitting of generated adversarial examples to the surrogate models. The proposed DFANet can be combined with existing black-box attack enhancement methods to achieve further improvements over them. Finally, based on our study, we contribute an adversarial face database as a benchmark to facilitate research on the robustness and defense of deep face recognition. 

The main contributions of our paper are as follows: 
\begin{itemize}
\item{We propose a dropout face attacking networks (DFANet) to enhance the transferability of adversarial attacks against deep face recognition. DFANet applies dropout layers to the surrogate model in each iteration of the adversarial examples generation process. As the number of iterations increases, the ensemble-like effects of the diverse generated dropout models can gradually improve the transferability of adversarial attacks.}

\item{We empirically demonstrate that momentum boosting method~\cite{dong2018boosting}, diverse input method~\cite{xie2019improving}, and model ensemble method~\cite{liu2016delving} are useful to improve the transferability of the feature-level attacks. They can be combined to provide a strong attacking baseline method for deep face recognition. Furthermore, the combination of the strong baseline with DFANet can obtain more effective black-box adversarial attacks towards the state-of-the-art face models with different training databases, loss functions and network architectures.}

\item{Based on the proposed DFANet, we generate the adversarial images from the well-known LFW database with visually imperceptible noise, which provides a new database, TALFW, to serve as a benchmark to evaluate the robustness of deep face models. With the same protocol as the LFW database, four state-of-the-art algorithms and four commercial APIs yield unpleasant performance on the TALFW database. The severe degradation clearly shows the vulnerability of deep face-recognition models even with massive training data.}
\end{itemize} 

The remainder of the paper is organized as follows. Section II briefly reviews the literature related to adversarial attacks, deep face recognition, and adversarial attacks against deep face models. In Section III, we first introduce the applicable strategies of transferable adversarial attacks against deep face recognition. Then, we present the proposed method and incorporate it into existing transferability enhancement methods to further improve adversarial transferability. In Section IV, we first demonstrate the superiority of the feature-level attack method and select it as the baseline method. Next, we evaluate the proposed method on deep face models with different training databases, loss functions and network architectures. Then, we provide an ablation study on hyperparameters and provide some interpretation of the intermediate generation process of adversarial examples for better comprehension of DFANet. We also compare the proposed method with other adversarial attack methods. Finally, we adopt the proposed method to the LFW database and build the TALFW database. Section V summarizes the conclusions.

\section{Related Works}
In this section, we briefly review the literature related to adversarial attacks, deep face recognition, and adversarial attacks against deep face models. 

\subsection{Adversarial Attacks}
\label{subsection:AdversarialAttacks}
Previous works have discovered that, with elaborate strategies, DCNNs can be easily fooled by test images with imperceptible noise~\cite{Szegedy14}. This type of image is called adversarial examples. The existence of adversarial examples has led to a variety of studies on adversarial defenses~\cite{Szegedy14,Kurakin17atscale,Tramer18}. 

Adversarial examples can be classified as white-box attacks~\cite{Szegedy14,goodfellow2014explaining,moosavi2016deepfool,Kurakin17,moosavi2017universal} and black-box attacks~\cite{papernot2017practical,dong2018boosting,inkawhich2019feature,huang2019enhancing}. White-box attacks assume that the attacker knows everything about the target models, including the network architectures, model parameters, training databases and even defensive methods. Black-box attacks assume that the attacker has no access to the target model. Some works broaden this restriction and assume that the attacker can obtain the output of the model (label or confidence score). Therefore, these works have developed a serious of query-based attack methods. In this paper, we assume that the attacker only has one chance to attack a face pair; therefore, they can only leverage the transferability of adversarial examples without any queries.

First, we introduce mainstream white-box attacks~\cite{Szegedy14,goodfellow2014explaining,moosavi2016deepfool,Kurakin17}. Szegedy \emph{et al}.\ \cite{Szegedy14} first find that DCNNs can be misled by a certain hardly perceptible perturbations generated with a box-constrained LBFGS method, despite this method is time-consuming. Goodfellow \emph{et al}.\ \cite{goodfellow2014explaining} propose a more time-saving and practical method, referred to as the fast gradient sign method (FGSM), which builds an adversarial example by performing one-step gradient updating along the direction of the sign of the gradient at each pixel. The fast target gradient sign method (FTGSM)~\cite{Kurakin17} modifies FGSM slightly to generate targeted attacks. FTGSM leads the model to misclassify a source image $x^{\left( s \right)}$ with label ${y^{(s)}}$ as another target category ${y^{(t)}}$, which is formulated as 
\begin{equation}\label{equ:FTGSM}
x_{adv} = x^{\left( s \right)}+\Delta x = {x^{(s)}} - \varepsilon sign({\nabla _{{x^{(s)}}}}J({x^{(s)}},{y^{(t)}})),\end{equation}where $J$ represents the cross-entropy loss, $\varepsilon$ limits the maximum deviation of the perturbation. Moreover, a straightforward method called the basic iterative method (BIM)~\cite{Kurakin17} extends the FGSM by applying it multiple times with a small step size and clip pixel values after each step to ensure the $L_\infty$ constraint. Compared with BIM~\cite{Kurakin17}, the version of targeted attacks is the iterative target gradient sign method (ITGSM)~\cite{Kurakin17}, formulated as
\begin{equation}\label{equ:ITGSM}\begin{gathered}
x_{adv,0} = {x^{(s)}},\hfill \\
x_{adv,N + 1} =  {C_{{x^{(s)}},\varepsilon }}(x_{adv,N} \!-\!  sign({\nabla _{x_{adv,N}}}J(x_{adv,N},{y^{(t)}}))), \hfill \\ 
\end{gathered}\end{equation}where the iteration can be chosen to be $\min (\varepsilon+4,1.25\varepsilon)$ and $C{_{x,\varepsilon }}(x') = \min (255,x + \varepsilon ,\max (0,x - \varepsilon ,x'))$. Iterative attacks can achieve a higher attack success rate than the fast attack method in a white-box setting, but perform worse when transferred to other models~\cite{Kurakin17}.

Compared with white-box attacks, transferable black-box attacks~\cite{liu2016delving, papernot2017practical,dong2018boosting,dong2019evading,xie2019improving,wu2020skip} are more practical in real-world situations and can severely hinder robustness. Researchers discover that attacking an ensemble of multiple models simultaneously~\cite{liu2016delving,dong2018boosting} can improve the transferability of adversarial images. Dong \emph{et al}.\ \cite{dong2018boosting} propose to integrate the momentum term into the attack process to stabilize the update directions and escape from poor local maxima, improving the transferability of iterative attacks. Furthermore, Dong \emph{et al}.\ \cite{dong2019evading} propose a translation-invariant
attack method that optimizes a perturbation over an ensemble of translated images. At the same time, Xie \emph{et al}.\ \cite{xie2019improving} propose to improve the transferability of adversarial examples by creating diverse input patterns. Wu \emph{et al}.\ \cite{wu2020skip} propose to improve the transferability by using more gradients from the skip connections rather than the residual modules according to a decay factor. These transferability enhancement methods increase the diversity and variability of the gradient~\cite{dong2018boosting,wu2020skip}, input images~\cite{dong2019evading,xie2019improving}, and surrogate models~\cite{liu2016delving}, so that adversarial examples are more transferable and generalizable, which significantly improve the applicability and practicability of black-box adversarial attacks. However, in these methods, the surrogate model are fixed, which have been trained before and will not change in the generation process of the adversarial examples. Therefore, our method fills this vacant area in the previous research, which changes the surrogate model in each iteration of the generation process in order to further increase the diversity of surrogate models.  

In addition to the image-specific attacks which generate different perturbations for each clean input images, Moosavi-Dezfooli \emph{et al}.\ \cite{moosavi2017universal} prove the existence of the image-agnostic adversarial attack, which can create a universal adversarial perturbation (UAP) for the images of whole database. To further improve the image-agnostic attacks, Mopuri \emph{et al}.\ \cite{mopuri2018generalizable} propose a data-free objective to learn perturbations that can adulterate the deep features extracted by multiple layers. Although this method demonstrates impressive fooling rates and surprising transferability across various deep models with different architectures, regularizers, and underlying tasks, UAP can only be applied in untargeted adversarial attacks.    

\subsection{Deep Face Recognition}
The success of deep face recognition can be mainly attributed to the large-scale training databases~\cite{Yi2014CASIA,Guo16MS,cao2018vggface2,wang2018devil}, effective training loss functions~\cite{sun2014deep,Schroff2015FaceNet,wen2016discriminative,liu2017sphereface,chen2017noisy,Wang2018CosFace,deng2019arcface} and advanced network architectures~\cite{6909616,sun2014deep,sun2015deepid3,parkhi2015deep,he2016deep,cao2018vggface2}. A comprehensive survey can be referred to~\cite{DBLP:journals/corr/abs-1804-06655}. 

Large scale training databases play an import role in deep face recognition, and we introduce four mainstream large scale databases here. CAISA-WebFace~\cite{Yi2014CASIA} is the first widely used large-scale training database in deep face recognition and contains 0.49M images of
10,575 celebrities. MS-Celeb-1M~\cite{Guo16MS} is the first publicly available million-scale training database. Considering the existence of label noise~\cite{wang2018devil}, the cleaned versions of MS-Celeb-1M are widely used in academia today. VGGFace2~\cite{cao2018vggface2} database has 3.31 million images of 9,131 identities and has large variations in pose, age, illumination, ethnicity and profession. IMDb-Face~\cite{wang2018devil} is a million-scale noise-controlled training database, containing 1.7M images of 59K identities, which is competitive as a training source despite its relatively smaller size.
 
It has been widely accepted that learning discriminative features is the key for open-set face recognition and the major focus in deep face recognition has become to training with effective loss functions. Some loss functions are based on Euclidean metric learning methods~\cite{sun2014deep, Schroff2015FaceNet}. Some loss functions modify softmax loss by incorporating weight or feature normalization~\cite{Ranjan17,wang2017normface,Zheng18ring}. Another powerful type of loss function is the large-margin softmax loss, mainly containing SphereFace~\cite{liu2017sphereface}, CosFace~\cite{Wang2018CosFace} and ArcFace~\cite{deng2019arcface}. Large-margin softmax loss functions significantly boost the performance, which seems to be the most effective loss functions in deep face recognition.

Network architectures have also shown significant gains in the deep face recognition literature. SphereNet~\cite{liu2017sphereface} adopts a 64-layer ResNet~\cite{he2016deep} network, supervised by an advanced large-margin loss function, achieving a 99.42\% accuracy on the LFW database. Adacos~\cite{zhang2019adacos} adopts an Inception-ResNet architecture~\cite{szegedy2017inception} and reports comparable results. ArcFace~\cite{deng2019arcface} develops ResNet~\cite{he2016deep} and squeeze-and-excitation Network (SENet)~\cite{hu2018squeeze} with an IR~\cite{deng2019arcface} block, which has a BN-Conv-BN-PReLU-Conv-BN structure, and achieves new state-of-art performance on several benchmarks. In addition, lightweight deep face models aiming at model compactness and computational efficiency have also attracted attention~\cite{deng2019lightweight}, where MobileNet~\cite{howard2017mobilenets} has been proven to be a successful attempt.

There are still many works insisting on designing more large-scale training databases, effective loss functions, and diverse network architectures. Such a wide variety of choices for deep face models increase the difficulty to generate transferable adversarial attacks in black-box settings.

\subsection{Attacks against Deep Face Models}
Face recognition systems can be applied in many safety critical domains. Therefore, it is important to understand the extent to which deep face models are subject to adversarial attacks. Previous works have already investigated several attacks~\cite{sharif2016accessorize,goswami2018unravelling,song2018attacks,dong2019efficient} and demonstrated the vulnerability of deep face models. 

Mahmood \emph{et al}.\ \cite{sharif2016accessorize} propose the first gradient-based attack method, which restricts the perturbation to eyeglasses and physically realizes impersonation and dodging attacks in deep face recognition. Then, the eyeglass adversarial attack is developed using generative methods~\cite{sharif2017adversarial}. The eyeglass adversarial attacks~\cite{sharif2016accessorize,sharif2017adversarial} not only investigate the vulnerability of deep face models, but also demonstrate the attacks can be physically realizable. The LOTS attack~\cite{rozsa2017lots} is proposed to form adversarial examples that mimic the deep features of the target. LOTS is the first work to launch feature-level attacks against deep face models, which shows similarities to the technique of Sabour \emph{et al}. \cite{sabour2016adversarial} in terms of directly adjusting internal feature representations. Dabouei \emph{et al}.\ \cite{dabouei2019fast} demonstrate that deep face models are vulnerable to geometrically-perturbed adversarial examples generated by a fast algorithm, which directly manipulates landmarks of the face images. 

To date, adversarial attacks against face models~\cite{sharif2016accessorize,sharif2017adversarial,rozsa2017lots,sabour2016adversarial,dabouei2019fast} have mainly focused on the white-box setting. Although the attack success rate in the white-box settings~\cite{sharif2016accessorize,rozsa2017lots} reaches almost 100\%, this setting is not practical in real-world situations due to the assumption that attackers have full access to the target models. Considering the impracticality of the white-box setting, Dong \emph{et al}.\ \cite{dong2019efficient} propose an evolutionary attack method for query-based adversarial attacks in the decision-based black-box settings. Despite the effectiveness in terms of the attack success rate, Query-based methods in the black-box setting~\cite{dong2019efficient} require a large number of queries, which can be relatively easy detected by the target system. 

Generative methods~\cite{Song18,deb2019advfaces} are also applied to generate adversarial examples to attack deep face models, which can be transferred to unseen models to some degree. Goswami \emph{et al}.\ \cite{goswami2019detecting} find that, in addition to sophisticated learning based attacks, face models can also be affected by image processing based adversarial attacks, such as grid based occlusion, most significant bit based noise, forehead and brow occlusion, eye region occlusion, and beard-like occlusion. These distortions are effective, easy to generate, and can be applied in black-box settings without knowing the information of target models. However, this type of attacks is visually noticeable and untargeted, which cannot be used to fool a face model into falsely predicting an face image as a specific identity.  

Therefore, in this work, we aim to study the transferable adversarial attacks against deep face models which can severely hinder the robustness, since this type of attack is launched in a fully black-box manner without any query feedback. We first investigate the gradient-based adversarial attack methods and the corresponding transferability enhancement methods. Then, we propose a new transferability enhancement method to generate adversarial examples by increasing the diversity of surrogate models. In addition, the proposed method is easy and stable for implementation. 

\section{Transferable Adversarial Attacks against Deep Face Models}
In this section, we first introduce the applicable feature-level adversarial attacks against deep face models. Then we propose the dropout face attacking networks (DFANet) to further increase the transferability of adversarial attacks, which can be combined with existing black-box attack enhancement methods, and achieve improvements over them.

\subsection{Adversarial Attacks against Deep Face Models}
\label{section:fromlabel2fea}
We start with some notations for the background of deep face models and the corresponding adversarial attacks. 

\textbf{Deep Face Models.} Let $D = \{ {x^{(i)}},{y^{(i)}}\}$ denote a labeled database, where ${x^{(i)}}$ and ${y^{(i)}} \in \{ 1,2,...,C\}$ ($C$ is the number of identities) denote an input image and the corresponding label, respectively. In the training process, a network with parameters $\theta_N$ is originally trained on a database $D$ by minimizing the cross-entropy loss\begin{equation}\label{loss} J({x^{(i)}},{y^{(i)}})=  - \frac{1}{N}\sum\limits_{i = 1}^N {\log p({y^{(i)}}|{x^{(i)}},\theta_N)},\end{equation}where $N$ is the mini-batch size. In the cross entropy loss function \begin{equation}\label{a:2} p({y^{(i)}}|{x^{(i)}},\theta_N) = \frac{{{e^{{{W_{y^{(i)}}}}^T{{x_i}} + {{b_{y^{(i)}}}}}}}}{{\sum\nolimits_{j = 1}^C {{e^{{{W_j}}^T{{x_i}} + {{b_j}}}}} }},\end{equation}${{x_i}} \in {\mathbb{R}^d}$ denotes the embedding feature of the $i$-th training image ${x^{(i)}}$, $y^{(i)}$ is the label of ${x^{(i)}}$, ${{{W}_{j}}} \in {\mathbb{R}^{\text{d}}}$ is the $j$-th column of the weight of the last fully connected layer, and ${b_j}\in {\mathbb{R}^{\text{C}}}$ is the bias. 

For a deep face model, in the training process, the network obtains an input face image ${x^{(i)}}$, and then outputs the probability $p({y^{(i)}}|{x^{(i)}},\theta_N)$ and the output label $l({x^{(i)}})$. In the testing process, we use the face model as a feature extractor, which means that we do not care about the softmax layer. We only extract the normalized embedding feature of images in the testing databases and make comparisons using distance metrics such as the normalized Euclidean distances. Some softmax-based loss functions~\cite{wang2017normface,liu2017sphereface,Wang2018CosFace,deng2019arcface} normalize the weight and embedding features, and incorporate the large margin, which improves the recognition performance but does not change the pipeline of deep face models.  

\textbf{Feature-level Attacks.} Given a deep face model and a face pair, denoted by $\{ {x^{(s)}},{x^{(t)}}\}$, we compare this face pair by calculating the distance between normalized deep representations $F(x^{(s)})$ and $F(x^{(t)})$. Note that in equation~\eqref{a:2}, ${{x_i}} \in {\mathbb{R}^d}$ denotes the embedding feature of the $i$-th training image ${x^{(i)}}$, while here, $F({x^{(i)}})$ is the deep feature after normalization, which is primarily used for distance comparison in deep face recognition. Ideally, the distance between features in a negative pair is larger than that in a positive pair. However, to explore the vulnerability of deep face models, we try to add imperceptible perturbation $\Delta {x}$ on one of the face images $x^{(s)}$ to generate an adversarial example $x_{adv} = x^{(s)} + \Delta {x}$ and deceive the face model. Therefore, for a positive face pair$\{ {x^{(s)}},{x^{(t)}}\}$, where ${y^{(s)}} = {y^{(t)}}$, the optimized objective can be formulated as 
\begin{equation}\label{fealoss1}
	\Delta {x} = \mathop {\arg \max }\limits_{\Delta {x}} \left\| {F(x^{(s)} + \Delta {x}) - F(x^{(t)})} \right\|_2, \left\| {\Delta {x}} \right\|_\infty  < \varepsilon.
	\end{equation} While for a negative face pair $\{ {x^{(s)}},{x^{(t)}}\}$, where ${y^{(s)}} \ne {y^{(t)}}$, the optimized objective is
\begin{equation}\label{fealoss2}
	\Delta {x} = \mathop {\arg \min }\limits_{\Delta {x}} \left\| {F(x^{(s)} + \Delta {x}) - F(x^{(t)})} \right\|_2, \left\| {\Delta {x}} \right\|_\infty  < \varepsilon,
	\end{equation}where $\varepsilon$ limits the maximum deviation of the perturbation. For computational efficiency, we adopt the optimized loss function \begin{equation}J\left( x^{\left( s \right)}+\Delta x, x^{\left( t \right)} \right) = \left\| F\left( x^{\left( s \right)}+\Delta x \right) -F\left( x^{\left( t \right)} \right) \right\|_2\end{equation} for negative pairs, and \begin{equation}J\left( x^{\left( s \right)}+\Delta x, x^{\left( t \right)} \right) = - \left\| F\left( x^{\left( s \right)}+\Delta x \right) -F\left( x^{\left( t \right)} \right) \right\|_2\end{equation} for positive pairs to form adversarial perturbations
	in a fast way, referred to as feature fast attack method (FFM)~\cite{zhong2019adversarial}
	\begin{equation}{x^{(s)}} + \Delta x = C{_{{x^{(s)}},\varepsilon }}({x^{(s)}} + sign(\nabla_{{x^{(s)}}}J( x^{\left( s \right)}, x^{\left( t \right)} ))),\end{equation}
	or in an iterative way, referred to as feature iterative attack method (FIM)~\cite{zhong2019adversarial}
\begin{equation}\begin{gathered}
	\label{euqa2}
	\Delta x_0 = 0,\hfill \\
	g_{N+1} = \nabla_{{x^{(s)}} + \Delta x_N}J\left( x^{\left( s \right)}+\Delta x_N, x^{\left( t \right)} \right),  \hfill \\
	{x^{(s)}} + \Delta x_{N + 1} = C{_{{x^{(s)}},\varepsilon }}({x^{(s)}} + \Delta x_N + sign(g_{N+1})), \hfill \\
	\end{gathered}\end{equation}where $C{_{x,\varepsilon }}(x') = \min (255,x + \varepsilon ,\max (0,x - \varepsilon ,x'))$, the iteration can be chosen heuristically $\min (\varepsilon+4,1.25\varepsilon)$. Note that in FFM and FIM, the loss functions $J$ are no longer the cross entropy loss as Equation (\ref{loss}), but the feature-level loss functions in Equation (\ref{fealoss1}) and Equation (\ref{fealoss2}).

\subsection{Dropout Face Attacking Networks (DFANet)}
To further improve the transferability of adversarial attacks, there are some previous works on transferability enhancement methods~\cite{liu2016delving,dong2018boosting,dong2019evading,xie2019improving,wu2020skip}, which we reviewed in Section~\ref{subsection:AdversarialAttacks}. The existing methods mainly improve the transferability of adversarial attacks by increasing the diversity and variability of the gradient~\cite{dong2018boosting,wu2020skip}, input images~\cite{dong2019evading,xie2019improving}, and surrogate models~\cite{liu2016delving} to prevent adversarial examples overfitting to surrogate models. However, in existing methods, a surrogate deep face model usually appears as a whole part, which can be expressed as $\theta_N$ in Equation~(\ref{loss}). The parameters $\theta_N$ always remain fixed in the forward propagation and backpropagation process of the adversarial example generation. In this work, we propose an easy and general method by increasing the diversity and variability of the surrogate model $\theta_N$ itself, even if surrogate models have already been trained well. The proposed method can be combined with many existing transferability enhancement methods and applied to a variety of convolutional neural network architectures. 

The basis of deep face models is deep convolutional neural networks (DCNNs) in which convolutional layers play an important role. Since the aim is to further improve the transferability by increasing the diversity and variability, we incorporate dropout~\cite{srivastava2014dropout} in the convolutional layers in the iterative steps of the generation process. Although dropout has brought great success to DCNNs, it is primarily used in the fully connected layers~\cite{simonyan2014very,szegedy2015going} by dropping units along with their connections during the training process to improve the performance of the trained model. We use dropout in the testing process when deep face models are used to generate adversarial examples to improve the transferability of adversarial examples. The dropout will increase the possible settings of the subnetworks and combines the possible subnetworks in the iterative steps. 

Specifically, for a face model composed of convolutional layers, given the output $o_i\in \mathbb{R}^n$ from the $i$-th convolutional layer, we first generate a mask $M_i\in \mathbb{R}^n$ where each element $m_i$ is independently sampled from a Bernoulli distribution with probability $p_d$:\begin{equation}\label{drop}m_i\sim Bernoulli\left( p_d \right), \qquad m_i\in M_i.\end{equation}Then, we use this mask to modify the output as $o_i = M_i \times o_i$, where $\times$ denotes the Hadamard product. We name this modified model as dropout face attacking networks (DFANet). To obtain the ensemble effect of the random sampling of the subnetworks, we incorporate the DFANet into the generation process of transferable adversarial examples. In the $N$-th iterative step of the adversarial example generation, we generate the mask ${M_i}^N$ for the output ${o_i}^N$ of the $i$-th convolutional layer using Equation~(\ref{drop}), then change the output to \begin{equation}{o_i}^N = {M_i}^N\times {o_i}^N\end{equation} in the forward propagation process. Accordingly, in the gradient backpropagation process, the same mask ${M_i}^N$ is used. In the $N$-th iterative step, for an input image $x$, the normalized deep representations can be denoted by $\tilde{F}^N(x)$.  

\begin{figure}[htbp]
	\center
	\includegraphics[width=0.9\linewidth]{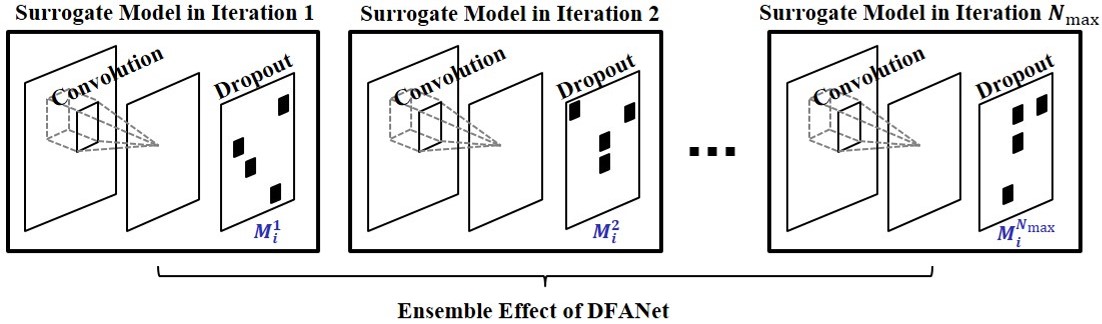}
	\caption{The surrogate model in the $N$-th step of adversarial examples generation is converted to another model by incorporating dropout layers. As the number of iterations increases, the ensemble-like effects of the diverse generated dropout models can gradually improve the transferability of adversarial attacks.}
	\label{fig:ensemble}
\end{figure}

The usage of DFANet can be understand as follows. The surrogate model in the $N$-th step of adversarial examples generation is converted to another model by incorporating dropout layers. As the number of iterations increases, the ensemble-like effects of the diverse generated dropout models can gradually improve the transferability of adversarial attacks, as shown in Fig.~\ref{fig:ensemble}.

\begin{algorithm}[htbp]
	\caption{DFANet-M-FIM}
	\label{alg:1}
	\LinesNumbered
	\KwIn{The face pair $\{ {x^{(s)}},{x^{(t)}}\}$, maximum deviation of perturbations $\varepsilon$, decay factor $\mu$, maximum iterative steps $N_{max}$, deep face model $F(\cdot)$, dropout probability $p_d$.}	
	\textbf{Initialize:} $\Delta x_0 = 0$, $g_0=0$, $N = 0$\;
	\While{step $N < N_{max}$}{
		\begin{small}Generate DFANet $\tilde{F}^N(x)$ of the $N$-th step (with $p_d$)\end{small}\;
		
		\begin{small}$J^N\left( x^{\left( s \right)}+\Delta x, x^{\left( t \right)} \right) = -\left\| \tilde{F}^N\left( x^{\left( s \right)}+\Delta x \right) -\tilde{F}^N\left( x^{\left( t \right)} \right) \right\|_2$\end{small}\;
		
		\begin{small}$g_{N+1}=\mu \cdot g_N+\frac{\nabla _{x^{\left( s \right)}+\Delta x_N}J\left( x^{\left( s \right)}+\Delta x_N,x^{\left( t \right)} \right)}{\lVert \nabla _{x^{\left( s \right)}+\Delta x_N}J\left( x^{\left( s \right)}+\Delta x_N,x^{\left( t \right)} \right) \rVert}$\end{small}\;
		
		\begin{small}${x^{(s)}} + \Delta x_{N+1} = C{_{{x^{(s)}},\varepsilon }}({x^{(s)}} + \Delta x_{N} + sign(g_{N+1}))$\end{small}\;	
	}
	\KwOut{Adversarial example $x_{adv}=x^{(s)}+\Delta x_{N_{max}}$.}	
\end{algorithm}

\begin{algorithm}[htbp]
	\caption{DFANet-DI-M-FIM}
	\label{alg:2}
	\LinesNumbered
	\KwIn{The face pair $\{ {x^{(s)}},{x^{(t)}}\}$, maximum deviation of perturbations $\varepsilon$, maximum iterative steps $N_{max}$, decay factor $\mu$, stochastic image transform function $T\left( x;p \right)$, dropout probability $p_d$.}	
	\textbf{Initialize:} $\Delta x_0 = 0$, $g_0=0$, $N = 0$\;
	\While{step $N < N_{max}$}{
		\begin{small}Generate DFANet $\tilde{F}^N(x)$ of the $N$-th step (with $p_d$)\end{small}\;
		
		\begin{small}$J^N\left( x^{\left( s \right)}+\Delta x, x^{\left( t \right)} \right) = -\left\| \tilde{F}^N\left( x^{\left( s \right)}+\Delta x \right) -\tilde{F}^N\left( x^{\left( t \right)} \right) \right\|_2$\end{small}\;
		
		\begin{small}$g_{N+1}=\mu \cdot g_N+\frac{\nabla _{x^{\left( s \right)}+\Delta x_N}J\left( T\left( x^{\left( s \right)}+\Delta x_N;p \right) ,x^{\left( t \right)} \right)}{\lVert \nabla _{x^{\left( s \right)}+\Delta x_N}J\left( T\left( x^{\left( s \right)}+\Delta x_N;p \right) ,x^{\left( t \right)} \right) \rVert}$\end{small}\;
		
		$x^{\left( s \right)}+\Delta x_{N+1}=C_{x^{\left( s \right)},\varepsilon}\text{(}x^{\left( s \right)}+\Delta x_N+sign\left( g_{N+1} \right)$\;	
	}
	\KwOut{Adversarial example $x_{adv}=x^{(s)}+\Delta x_{N_{max}}$.}	
\end{algorithm}

\begin{algorithm}[htbp]
	\caption{DFANet-E-DI-M-FIM}
	\label{alg:3}
	\LinesNumbered
	\KwIn{The face pair $\{ {x^{(s)}},{x^{(t)}}\}$, maximum deviation of perturbations $\varepsilon$, maximum iterative steps $N_{max}$, decay factor $\mu$, stochastic image transform function $T\left( x;p \right)$,  dropout probability $p_d$, ensemble weight $w_k$ ($w_k\geqslant 0$, $K$ deep face models $F_k(\cdot)$, $\sum\nolimits_{k=1}^K{w_k}=1$).}	
	\textbf{Initialize:} $\Delta x_0 = 0$, $g_0=0$, $N = 0$\;
	\While{step $N < N_{max}$}{
		\While{model $k<K$}{Generate DFANet $\tilde{F}_k^N(x)$ of the $N$-th step and the $k$-th model (with $p_d$)\;}
		\begin{small}$J_k^N\left( x^{\left( s \right)}+\Delta x, x^{\left( t \right)} \right) = -\left\| \tilde{F}_k^N\left( x^{\left( s \right)}+\Delta x \right) -\tilde{F}_k^N\left( x^{\left( t \right)} \right) \right\|_2$\end{small}\;
		
		\begin{small}$g_{N+1}=\mu \cdot g_N+\frac{\sum\nolimits_{k=1}^K{w_k}\nabla _{x^{\left( s \right)}+\Delta x_N}J_k^N\left( T\left( x^{\left( s \right)}+\Delta x_N;p \right) ,x^{\left( t \right)} \right)}{\lVert \sum\nolimits_{k=1}^K{w_k}\nabla _{x^{\left( s \right)}+\Delta x_N}J_k^N\left( T\left( x^{\left( s \right)}+\Delta x_N;p \right) ,x^{\left( t \right)} \right) \rVert}$\end{small}\;
		
		$x^{\left( s \right)}+\Delta x_{N+1}=C_{x^{\left( s \right)},\varepsilon}\text{(}x^{\left( s \right)}+\Delta x_N+sign\left( g_{N+1} \right)$\;	
	}
	\KwOut{Adversarial example $x_{adv}=x^{(s)}+\Delta x_{N_{max}}$.}	
\end{algorithm}

DFANet can be applied to the generation of FIM (by Equation~(\ref{euqa2})) and combined with transferability enhancement methods like momentum boosting method~\cite{dong2018boosting}, diverse input method~\cite{xie2019improving}, and model ensemble method~\cite{liu2016delving}. To explain the generation process more clearly, we detail these combinations of methods in turn. For convenience, we do not detail the combination of DFANet and FIM here. We start with DFANet-M-FIM, which is the combination of DFANet, momentum boosting method, and FIM. DFANet-M-FIM is shown in Algorithm~\ref{alg:1}, where the integrated momentum stabilizes the update directions and prevents the optimization from dropping into poor local maxima. Specifically, $g_{N+1}$ gathers the gradients of the first $N+1$ iterations with a decay factor $\mu$. The combination of DFANet, diverse input method, momentum boosting method, and FIM is referred to as DFANet-DI-M-FIM and is shown in Algorithm~\ref{alg:2}. In Algorithm~\ref{alg:2}, $T\left( x;p \right)$ is the stochastic image transform function (\emph{e.g}., resizing, cropping and rotating), creating diverse input patterns~\cite{xie2019improving} to improve the transferability:
\begin{equation}\label{euqa_input}
T\left( x;p \right) =\begin{cases}
T\left( x \right)&		with\,\,probability\,\,p\\
x&		with\,\,probability\,\,1-p\\
\end{cases}.\end{equation}Finally, we describe the most complex one: the combination of DFANet, model ensemble method, diverse input method, the momentum boosting method, and FIM, referred to as DFANet-E-DI-M-FIM. Algorithm~\ref{alg:3} details DFANet-E-DI-M-FIM, where instead of optimizing a single face model $J$, we apply model ensemble method by attacking $K$ models. In addition, the feature loss functions in Line 4 of Algorithm~\ref{alg:1}, Algorithm~\ref{alg:2} and Line 6 of Algorithm~\ref{alg:3} can be replaced as those for positive pairs, but for convenience, we input negative pairs as examples here.

\section{Experiments}
In this section, we first explore applicable strategies of adversarial attacks against deep face models, from label-level attacks to feature-level attacks to select a baseline method for further study. Then we conduct experiments on state-of-the-art face models with various training databases, training loss functions and network architectures to evaluate the proposed DFANet. Next, we compare the proposed DFANet with other attack methods. Finally, we apply the DFANet to the LFW database and generate new face pairs with adversarial perturbations, called the TALFW database. We use the LFW and TALFW databases to evaluate the robustness of mainstream open-sourced deep face models, commercial APIs and defensive methods.

\begin{figure}[ht]
	\center
	\includegraphics[width=1\linewidth]{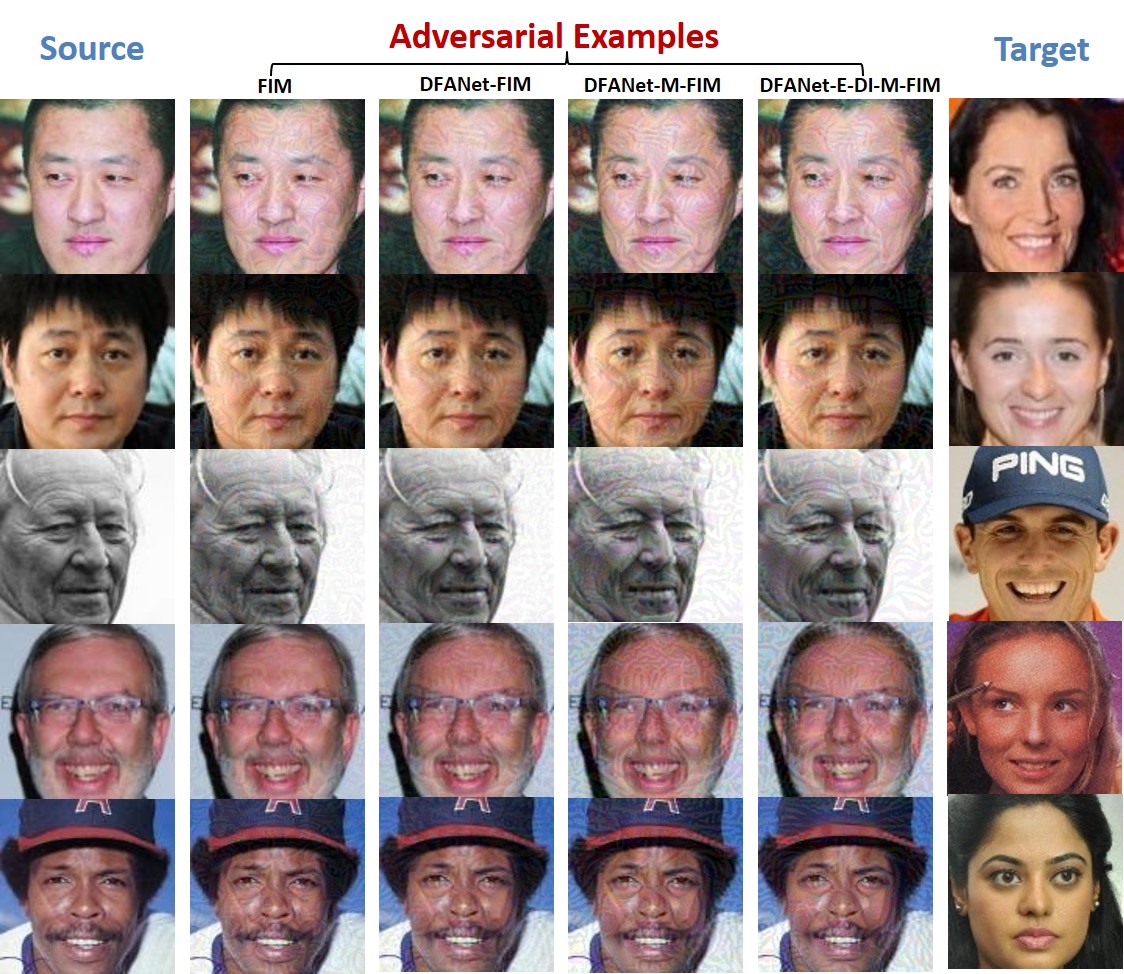}
	\caption{The first column shows some of the source images and the last column lists some of the target images. The second to the fifth column shows the corresponding adversarial examples generated by FIM, DFANet-FIM, DFANet-M-FIM, and DFANet-E-DI-M-FIM under maximum $L_{\infty}$ perturbation $\varepsilon=10$ with respect to pixel values in $\left[ 0,255 \right]$. We generate a total of 10,000 pairs from the 100 source images and 100 target images, which were originally judged as negative pairs.}
	\label{eval_pairs}
\end{figure}

\subsection{Experimental Settings and Evaluation Protocol}
We use target adversarial examples to evaluate transferability since targeted attacks are more difficult to transfer between deep models~\cite{liu2016delving}. (1) First, we selected 100 source images from MS-Celeb-1M~\cite{Guo16MS} and 100 target images from VGGFace2~\cite{cao2018vggface2} to generate 10,000 pairs, which are originally judged as negative pairs. Some of the source and target images are shown in Fig.~\ref{eval_pairs}, where the first column shows the source images, the second to the fifth column shows the corresponding adversarial examples generated by DFANet-FIM, DFANet-M-FIM, and DFANet-E-DI-M-FIM under maximum $L_{\infty}$ perturbation $\varepsilon=10$ with respect to pixel values in $\left[ 0,255 \right]$, and the last column shows target images. Therefore, the goal of the attack is to generate adversarial examples from the source images to be disguised as target ones. Specifically, the goal is to obtain face embedding representations of source images closer to those of target images than the distance threshold of a face recognition system. (2) Next, we define the threshold $t_m$ of a face model $m$. Using 6,000 face pairs from the LFW database~\cite{LFWTech}, we compute the Euclidean distance of normalized deep features to obtain ROC curves. Then we identify distance thresholds for judging whether a pair is positive or negative. Since we would like to compare the adversarial robustness of the trained models with real-world applications, we define the distance threshold for attacking (or distinguishing a positive and a negative pair) to have a low false acceptance rate (FAR = $1e-3$). An attack is defined as a success (hit) if the embedding distance between the source image and target is less than the threshold. We use the average hit rate of the $N_p=10,000$ face pairs to report the transferable attack success (hit) rate:\begin{equation}Hit\,Rate\,\left(\%\right)=100\times\frac{1}{N_p}\sum_{i=1}^{N_p}{ 1\left[ \left\| F\left( x_i^{\left( s \right)} \right) -F\left( x_i^{\left( t \right)} \right) \right\| _2<t_m \right]}.\end{equation}The higher the hit rate is, the stronger the transferability of the attack.

To simulate potential scenarios, we aim to increase the difference between the source and target models by attacking deep face models with different training databases, training loss functions, and network architectures. The first setting is attacks between models with different training databases. Specifically, we use the four mainstream training databases in the deep face recognition literature, namely, CAISA-WebFace~\cite{Yi2014CASIA}, MS-Celeb-1M~\cite{Guo16MS}, VGGFace2~\cite{cao2018vggface2} and IMDb-Face~\cite{wang2018devil}, to train a modified version~\cite{deng2019arcface} of a ResNet-50~\cite{he2016deep} model supervised by softmax loss. The statistics for the four training databases can be referred to the Appendix. The second setting is attacks between models with different training loss functions. We use the softmax loss, triplet loss~\cite{Schroff2015FaceNet}, CosFace~\cite{Wang2018CosFace} and ArcFace~\cite{deng2019arcface} in the experiment. Then the third setting is attacks between models with different network architectures. We use the modified version~\cite{deng2019arcface} of ResNet~\cite{he2016deep}, the modified version~\cite{deng2019arcface} of squeeze-and-excitation Network (SENet)~\cite{hu2018squeeze}, MobileNet~\cite{howard2017mobilenets}, and Inception-ResNet~\cite{szegedy2017inception}. To keep it simple, we refer to the modified version~\cite{deng2019arcface} of ResNet and SENet with 50 layers as ResNet-50 and SENet-50, which both use IR blocks~\cite{deng2019arcface} with a BN-Conv-BN-PReLU-Conv-BN structure. The recognition performance of these models can be referred to the Appendix.

\subsection{Experiments for the Baseline Method}
To explore adversarial attacks against deep face models and find a baseline method for further study on transferability enhancement methods, we compare the attack performance of applicable adversarial attacks against deep face models including label-level and feature-level attacks.

We use FTGSM~\cite{Kurakin17} and ITGSM~\cite{Kurakin17} to generate label-level adversarial examples and FFM~\cite{zhong2019adversarial} and FIM~\cite{zhong2019adversarial} to generate feature-level adversarial examples. Since we introduced the feature level in detail earlier, let us clarify something noticeable in the label-level attacks against deep face recognition. In label-level attacks, several gradient-based generative strategies, including FGSM~\cite{goodfellow2014explaining}, FTGSM~\cite{Kurakin17}, BIM~\cite{Kurakin17}, and ITGSM~\cite{Kurakin17} have been proposed in the literature. However, in the deep face recognition, test identities are usually different from those in the training databases. We actually cannot always use ${y^{(s)}}$ and ${y^{(t)}}$ for comparison. Therefore, first, we forward propagate the input images and obtain the output labels $l(x^{(s)})$ and $l(x^{(t)})$. In the following label-level adversarial attacks, we use $l(x^{(s)})$ and $l(x^{(t)})$ to replace ${y^{(s)}}$ and ${y^{(t)}}$ in FTGSM (Equation~(\ref{equ:FTGSM})) and ITGSM (Equation~(\ref{equ:ITGSM})). Specifically, the maximum deviation of perturbations $\varepsilon$ is set to 10. Correspondingly, the number of iterations for ITGSM and FIM are chosen to be $\min (\varepsilon+4,1.25\varepsilon)\approx 13$. We use four aforementioned deep face models trained on different training databases. 

The experimental results are shown in Table~\ref{tabel:Label2FEA}. In terms of the transferability of the black-box attacks, feature-level attacks (FFM and FIM) are much more effective than label-level attacks (FTGSM and ITGSM) at the same constraint level. In fact, the lack of accurate labels and the large number of categories not only increase the difficulty of label-level attacks but also reduce the similarity between the source and target models. Instead, the embedding space of deep face models share more similarities, which benefits the transferability of adversarial attacks. These properties may explain why feature-level attacks are more transferable than label-level attacks to some extent. In addition, for both label-level and feature-level attacks, iterative methods (ITGSM, FIM) are much more effective than fast methods (FTGSM, FFM) at the same constraint level. Taken together, we choose FIM as the baseline method in the following experiments.

\begin{table}
	\newcommand{\tabincell}[2]{\begin{tabular}{@{}#1@{}}#2\end{tabular}}
	\renewcommand\arraystretch{1.0}
	\begin{center}
		\caption{Transferable adversarial success (hit) rate of label-level adversarial examples generated by FTGSM and ITGSM, and feature-level adversarial examples generated by FFM and FIM between four deep face models of different training databases(CAISA-WebFace~\cite{Yi2014CASIA}, MS-Celeb-1M~\cite{Guo16MS}, VGGFace2~\cite{cao2018vggface2} and IMDb-Face~\cite{wang2018devil}) and supervised by the softmax loss. }
		\label{tabel:Label2FEA}
		\scalebox{0.845}{
			\begin{tabular}{|c|c|c|c|c|c|}
				\hline
				Method                 & Src $\downarrow$ | Tar  $\rightarrow$          & WebFace & IMDB-Face&VGGFace2&MS1M\\ \hline\hline
				\multirow{4}{*}{FTGSM~\cite{Kurakin17}} & WebFace   & /                   & 1.85\%                & 1.18\%               & 1.37\%           \\ \cline{2-6} 
				& IMDB-Face & 2.84\%              & /                     & 1.87\%               & 2.42\%           \\ \cline{2-6} 
				& VGGFace2& 13.16\%             & 12.80\%               & /                    & 10.74\%          \\ \cline{2-6} 
				& MS1M& 5.80\%              & 6.87\%                & 4.94\%               & /                \\ \hline\hline
				\multirow{4}{*}{ITGSM~\cite{Kurakin17}} 
				&WebFace
				& / & 2.00\%& 1.25\%& 1.17\% \\ \cline{2-6} 
				& IMDB-Face 
				& 4.03\% & / & 2.34\%& 2.27\% \\ \cline{2-6} 
				&VGGFace2
				& 31.92\% & 28.00\%& /& 16.01\% \\ \cline{2-6} 
				& MS1M 
				& 15.20\%& 18.34\%& 11.65\%& / \\ \hline\hline
				\multirow{4}{*}{FFM~\cite{zhong2019adversarial}}   &WebFace& /                   & 9.44\%                & 7.47\%               & 6.11\%           \\ \cline{2-6} 
				& IMDB-Face & 15.76\%             & /                     & 9.99\%               & 11.50\%          \\ \cline{2-6} 
				&VGGFace2& 23.60\%             & 19.11\%               & /                    & 15.18\%          \\ \cline{2-6} 
				& MS1M& 14.05\%             & 14.71\%               & 10.54\%              & /                \\ \hline\hline
				\multirow{4}{*}{FIM~\cite{zhong2019adversarial}}  
				&WebFace
				& /& 11.10\%& 7.29\%& 4.95\%\\ \cline{2-6} 
				& IMDB-Face 
				& 30.90\%& /& 15.67\%& 12.71\%\\ \cline{2-6} 
				&VGG2
				& 65.95\%& 53.90\%& /& 32.24\%\\ \cline{2-6} 
				& MS1M
				& 36.43\%& 37.57\%& 25.19\%& /\\ \hline
			\end{tabular}
		} 
	\end{center}
\end{table}

\subsection{Strong Baseline and DFANet}
As Table~\ref{tabel:Label2FEA} shows, with the basic iterative label-level and feature-level method, it is still hard to guarantee the consistent success of transferable adversarial attacks against deep face recognition. Therefore, we intend to incorporate the transferability enhancement method~\cite{liu2016delving,dong2018boosting,xie2019improving} into the baseline method, FIM, to serve as a strong baseline. In addition, we propose DFANet to further improve the transferability. 
 
Apart from deep face models trained on different training databases, we also experiment using deep face models trained with different loss functions and network architectures. We first extend attack methods from FIM to DFANet-FIM, from M-FIM to DFANet-M-FIM, and from DI-M-FIM to DFANet-DI-M-FIM. Specifically, the maximum deviation of perturbations $\varepsilon$ is set to 10. For integrating momentum terms, the decay factor $\mu$ is set to 1. For the method of incorporating diverse input patterns, we introduce several transformations including translation, rotation and scaling. The transformation probability $p$ is set to 1 as in the original paper to reach the maximum transferability. In addition, we also try adding Gaussian noise but it has little effect. For the proposed DFANet, the maximum iterative step $N_{max}$ is set to 1,500, the drop rate $p_d$ of ResNet-50 and SENet-50 model is set to 0.1, $p_d$ of MobileNet model is set to 0.025 and $p_d$ of Inception-ResNet model is set to 0.05. We will discuss the effects of hyperparameters $p_d$ and $N_{max}$ in the following experiments. The experimental results from FIM to DFANet-FIM, from M-FIM to DFANet-M-FIM, and from DI-M-FIM to DFANet-DI-M-FIM are listed in Table~\ref{table:DFANet}, Table~\ref{table:DFANet2} and Table~\ref{table:DFANet3}. We can see a constant improvement, from FIM to DFANet-FIM, from M-FIM to DFANet-M-FIM, and from DI-M-FIM to DFANet-DI-M-FIM in terms of the hit rate between deep face models trained with different training databases, loss functions and network architectures. Compared with the strong baseline DI-M-FIM, DFANet-DI-M-FIM still achieves significant improvement. In addition, compared with the initial FIM, most of the successful hit rates of adversarial examples generated by DFANet-DI-M-FIM between the four models have been improved to approximately 90\%.

\begin{table}
	\newcommand{\tabincell}[2]{\begin{tabular}{@{}#1@{}}#2\end{tabular}}
	\renewcommand\arraystretch{1.0}
	\begin{center}
		\caption{Evaluation transferability enhancement methods from FIM to DFANet-FIM, from M-FIM to DFANet-M-FIM, and from DI-M-FIM to DFANet-DI-M-FIM. The methods are evaluated by success (hit) rate of adversarial examples between four deep face models (the ResNet-50 model trained on CAISA-WebFace~\cite{Yi2014CASIA}, MS-Celeb-1M~\cite{Guo16MS}, VGGFace2~\cite{cao2018vggface2} and IMDb-Face~\cite{wang2018devil} and supervised by the softmax function.)}
		\label{table:DFANet}
		\scalebox{0.852}{
			\begin{tabular}{|c|c|c|c|c|c|}
				\hline
				Method                 & Src $\downarrow$ | Tar  $\rightarrow$          & WebFace & IMDB-Face&VGGFace2&MS1M\\ \hline\hline
				\multirow{4}{*}{FIM~\cite{zhong2019adversarial}}&WebFace
				& /& 11.10\%& 7.29\%& 4.95\%\\ \cline{2-6} 
				& IMDB-Face 
				& 30.90\%& /& 15.67\%& 12.71\%\\ \cline{2-6} 
				&VGG2
				& 65.95\%& 53.90\%& /& 32.24\%\\ \cline{2-6} 
				& MS1M
				& 36.43\%& 37.57\%& 25.19\%& /\\ \hline\hline
				\multirow{4}{*}{\tabincell{c}{DFANet-\\FIM}}   
				&WebFace
				& /&46.81\%&39.63\%&25.82\% \\ \cline{2-6} 
				& IMDB-Face
				&81.39\%& / &67.31\% &58.70\% \\ \cline{2-6} 
				&VGGFace2
				&91.91\%&88.01\%& / &75.91\% \\ \cline{2-6} 
				& MS1M
				&64.38\%&67.06\%&54.56\%& / \\ \hline\hline\hline
				\multirow{4}{*}{\tabincell{c}{M-FIM\\ \cite{dong2018boosting}\cite{zhong2019adversarial}}} 
				&WebFace
				& /& 26.81\%& 22.14\%& 14.52\% \\ \cline{2-6} 
				& IMDB-Face
				& 49.86\%& / & 36.57\% & 31.24\% \\ \cline{2-6} 
				&VGGFace2
				& 66.57\%& 55.76\%& / & 42.81\% \\ \cline{2-6} 
				& MS1M
				& 43.43\%& 43.52\%& 36.43\%& / \\ \hline\hline
				\multirow{4}{*}{\tabincell{c}{DFANet-\\M-FIM}}   
				&WebFace
				& /&80.73\%&80.16\%&58.85\% \\ \cline{2-6} 
				& IMDB-Face
				&95.18\%& / &92.14\% &85.31\% \\ \cline{2-6} 
				&VGGFace2
				&96.08\%&93.23\%& / &87.30\% \\ \cline{2-6} 
				& MS1M
				&84.66\%&86.40\%&82.78\%& / \\ \hline\hline\hline
				\multirow{4}{*}{\tabincell{c}{DI-M-FIM\\
						\cite{xie2019improving}\cite{dong2018boosting}\cite{zhong2019adversarial}}}   
				&WebFace
				& /& 73.92\%& 74.05\%& 50.22\% \\ \cline{2-6} 
				& IMDB-Face
				& 88.82\%& / & 84.01\% & 71.53\% \\ \cline{2-6} 
				&VGGFace2
				& 91.86\%& 88.31\%& / & 78.31\% \\ \cline{2-6} 
				& MS1M
				& 80.15\%& 82.63\%& 80.72\%& / \\ \hline\hline
				\multirow{4}{*}{\tabincell{c}{DFANet-\\DI-M-FIM}}   
				&WebFace
				& /& 85.57\%& 84.52\%& 64.00\% \\ \cline{2-6} 
				& IMDB
				& 96.10\%& / & 92.90\% & 84.86\% \\ \cline{2-6} 
				&VGGFace2
				& 96.76\%& 93.48\%& / & 86.29\% \\ \cline{2-6} 
				& MS1M
				& 92.19\%& 92.90\%& 90.05\%& / \\ \hline
			\end{tabular}
		} 
	\end{center}
\end{table}

\begin{table}
	\newcommand{\tabincell}[2]{\begin{tabular}{@{}#1@{}}#2\end{tabular}}
	\renewcommand\arraystretch{1.0}
	\begin{center}
		\caption{Evaluation transferability enhancement methods from FIM to DFANet-FIM, from M-FIM to DFANet-M-FIM, and from DI-M-FIM to DFANet-DI-M-FIM. The methods are evaluated by the success (hit) rate of adversarial examples between four deep face models (the ResNet-50 model trained on CAISA-WebFace~\cite{Yi2014CASIA} supervised by the softmax loss, triplet loss~\cite{Schroff2015FaceNet}, CosFace~\cite{Wang2018CosFace} and ArcFace~\cite{deng2019arcface}).}
		\label{table:DFANet2}
		\scalebox{0.852}{
			\begin{tabular}{|c|c|c|c|c|c|}
				\hline
				Method& Src $\downarrow$ | Tar  $\rightarrow$
				& Softmax&Triplet Loss&CosFace&ArcFace\\ \hline\hline
				\multirow{4}{*}{FIM~\cite{zhong2019adversarial}}
				&Softmax
				& /&9.92\%&20.69\%&24.36\%\\ \cline{2-6} 
				&Triplet Loss
				&27.17\%& /&57.82\%&61.32\%\\ \cline{2-6} 
				&CosFace
				&11.78\%&14.13\%& /&31.56\%\\ \cline{2-6} 
				&ArcFace
				&6.43\%&7.93\%&16.89\%& /\\ \hline\hline
				
				\multirow{4}{*}{\tabincell{c}{DFANet-\\FIM}}   
				&Softmax
				& /&42.15\%&63.19\%&67.90\%\\ \cline{2-6} 
				&Triplet Loss
				&82.29\%& /&94.23\%&94.84\%\\ \cline{2-6} 
				&CosFace
				&53.90\%&40.92\%& /&80.08\%\\ \cline{2-6} 
				&ArcFace
				&46.71\%&34.53\%&72.53\%& /\\ \hline\hline\hline
				
				\multirow{4}{*}{\tabincell{c}{M-FIM\\ \cite{dong2018boosting}\cite{zhong2019adversarial}}} 
				&Softmax
				& /&29.46\%&38.95\%&44.15\%\\ \cline{2-6} 
				&Triplet Loss
				&36.31\%& /&67.73\%&73.09\%\\ \cline{2-6} 
				&CosFace
				&21.08\%&32.89\%& /&50.94\%\\ \cline{2-6} 
				&ArcFace
				&15.18\%&24.13\%&36.39\%& /\\ \hline\hline
				
				\multirow{4}{*}{\tabincell{c}{DFANet-\\M-FIM}}   
				&Softmax
				& /&80.16\%&89.71\%&91.01\%\\ \cline{2-6} 
				&Triplet Loss
				&89.85\%& /&97.56\%&97.81\%\\ \cline{2-6} 
				&CosFace
				&72.69\%&70.28\%& /&92.55\%\\ \cline{2-6} 
				&ArcFace
				&77.36\%&73.18\%&93.32\%& /\\ \hline\hline\hline
				
								\multirow{4}{*}{\tabincell{c}{DI-M-FIM\\
						\cite{xie2019improving}\cite{dong2018boosting}\cite{zhong2019adversarial}}}    
				&Softmax
				& /&73.39\%&86.09\%&87.22\%\\ \cline{2-6} 
				&Triplet Loss
				&84.94\%& /&95.55\%&96.67\%\\ \cline{2-6} 
				&CosFace
				&74.36\%&81.79\%& /&93.21\%\\ \cline{2-6} 
				&ArcFace
				&76.09\%&84.20\%&93.94\%& /\\ \hline\hline
				
				\multirow{4}{*}{\tabincell{c}{DFANet-\\DI-M-FIM}}   
				&Softmax
				&/&82.50\%&91.77\%&92.55\%\\ \cline{2-6} 
				&Triplet Loss
				&91.81\%& /&97.70\%&98.34\%\\ \cline{2-6} 
				&CosFace
				&82.32\%&88.41\%& /&95.72\%\\ \cline{2-6} 
				&ArcFace
				&82.59\%&88.76\%&95.59\%& /\\ \hline
			\end{tabular}
		} 
	\end{center}
\end{table}

Furthermore, to evaluate the ability of DFANet to enhance the model ensemble method, we generate adversarial examples using E-DI-M-FIM and DFANet-E-DI-M-FIM, and apply this attacks towards a commercial API, Face++~\cite{Face++}. Specifically, we generate three groups of attacks: the first group combines three ResNet-50 models trained on CAISA-WebFace and supervised by the softmax loss, CosFace, and ArcFace, and the second and third groups use the VGGFace2 and MS1M models respectively, with these three loss functions. The other parameters are set as above. We input the generated face pair to Face++, and the API returns both the similarity of this pair and the similarity threshold with FAR at 1e-3, 1e-4 and 1e-5. With the output information as aforementioned, we can calculate the success (hit) rate as before. We list the results in Table~\ref{DFANet_api}, from which we can see that with the combination of DFANet, the transferability of adversarial attacks generated from E-DI-M-FIM can be improved further. Above all, the extensive experiments on the face models with different training databases, loss functions and network architectures, shown in Table~\ref{table:DFANet}, Table~\ref{table:DFANet2} and Table~\ref{DFANet_api}, convincingly demonstrate the effectiveness of DFANet based on the strong baseline. 

\begin{table}
	\newcommand{\tabincell}[2]{\begin{tabular}{@{}#1@{}}#2\end{tabular}}
	\renewcommand\arraystretch{1.0}
	\begin{center}
		\caption{Evaluation transferability enhancement methods from FIM to DFANet-FIM, from M-FIM to DFANet-M-FIM, and from DI-M-FIM to DFANet-DI-M-FIM. The method are evaluated by the success (hit) rate of the adversarial examples between four deep face models (trained on CAISA-WebFace~\cite{Yi2014CASIA} use the softmax loss with different network architectures including ResNet-50~\cite{he2016deep}, SENet-50~\cite{hu2018squeeze}, MobileNet~\cite{howard2017mobilenets}, and Inception-ResNet~\cite{szegedy2017inception}).}
		\label{table:DFANet3}
		\scalebox{0.805}{
			\begin{tabular}{|c|c|c|c|c|c|}
				\hline
				Method& Src $\downarrow$ | Tar  $\rightarrow$
				&ResNet-50&SENet-50&MobileNet&Incep-ResNet\\ \hline\hline
				\multirow{4}{*}{FIM~\cite{zhong2019adversarial}}
				&ResNet-50
				& /&39.91\%&13.46\%&11.62\%\\ \cline{2-6} 
				&SENet-50
				&52.13\%& /&18.49\%&15.32\%\\ \cline{2-6} 
				&MobileNet
				&12.92\%&12.76\%& /&4.19\%\\ \cline{2-6} 
				&Incep-ResNet
				&18.54\%&17.84\%&9.32\%& /\\ \hline\hline
				
				\multirow{4}{*}{\tabincell{c}{DFANet-\\FIM}}   
				&ResNet-50
				& /&90.76\%&67.51\%&54.20\%\\ \cline{2-6} 
				&SENet-50
				&88.62\%& /&49.59\%&36.62\%\\ \cline{2-6} 
				&MobileNet
				&36.28\%&36.38\%& /&11.74\%\\ \cline{2-6} 
				&Incep-ResNet
				&41.50\%&40.66\%&23.57\%& /\\ \hline\hline\hline
				
				\multirow{4}{*}{\tabincell{c}{M-FIM\\ \cite{dong2018boosting}\cite{zhong2019adversarial}}} 
				&ResNet-50
				& /&65.43\%&29.77\%&27.68\%\\ \cline{2-6} 
				&SENet-50
				&77.48\%& /&37.57\%&34.46\%\\ \cline{2-6} 
				&MobileNet
				&32.82\%&33.50\%& /&13.52\%\\ \cline{2-6} 
				&Incep-ResNet
				&35.89\%&36.34\%&20.18\%& /\\ \hline\hline
				
				\multirow{4}{*}{\tabincell{c}{DFANet-\\M-FIM}}   
				&ResNet-50
				& /&99.17\%&92.41\%&86.14\%\\ \cline{2-6} 
				&SENet-50
				&98.24\%& /&78.58\%&67.79\%\\ \cline{2-6} 
				&MobileNet
				&62.14\%&64.45\%& /&27.78\%\\ \cline{2-6} 
				&Incep-ResNet
				&62.15\%&62.92\%&41.59\%& /\\ \hline\hline\hline
				
				\multirow{4}{*}{\tabincell{c}{DI-M-FIM\\
						\cite{xie2019improving}\cite{dong2018boosting}\cite{zhong2019adversarial}}}    
				&ResNet-50
				& /&98.76\%&84.16\%&78.14\%\\ \cline{2-6} 
				&SENet-50
				&99.02\%& /&85.56\%&77.95\%\\ \cline{2-6} 
				&MobileNet
				&85.94\%&87.30\%& /&54.24\%\\ \cline{2-6} 
				&Incep-ResNet
				&83.61\%&83.59\%&65.03\%& /\\ \hline\hline
				
				\multirow{4}{*}{\tabincell{c}{DFANet-\\DI-M-FIM}}   
				&ResNet-50
				& /&99.68\%&92.77\%&87.61\%\\ \cline{2-6} 
				&SENet-50
				&99.53\%& /&94.22\%&88.41\%\\ \cline{2-6} 
				&MobileNet
				&92.07\%&92.01\%& /&64.80\%\\ \cline{2-6} 
				&Incep-ResNet
				&88.98\%&88.56\%&73.44\%& /\\ \hline
			\end{tabular}
		} 
	\end{center}
\end{table}

\begin{table}
	\newcommand{\tabincell}[2]{\begin{tabular}{@{}#1@{}}#2\end{tabular}}
	\renewcommand\arraystretch{1.0}
	\begin{center}
		\caption{Evaluation transferability enhancement methods from E-DI-M-FIM to DFANet-E-DI-M-FIM. The methods are evaluated by the transferable adversarial success (hit) rate of adversarial examples on Face++~\cite{Face++}.}
		\label{DFANet_api}
		\scalebox{0.88}{
			\begin{tabular}{|c|c|c|c|c|}
				\hline
				\multirow{2}{*}{Method} & \multirow{2}{*}{Surrogate Models} & \multicolumn{3}{c|}{Hit Rate@FAR} \\ \cline{3-5} 
				&                                   & 1e-5      & 1e-4      & 1e-3      \\ \hline\hline
			
				\tabincell{c}{E-DI-M-FIM\\\cite{liu2016delving}\cite{xie2019improving}\cite{dong2018boosting}\cite{zhong2019adversarial}}&
				\tabincell{c}{ResNet-50,WebFace\\(Softmax,CosFace,ArcFace)}
				&51.90\%&67.20\%&80.50\%\\ \hline\hline
				
				\tabincell{c}{DFANet-\\E-DI-M-FIM}&
				\tabincell{c}{ResNet-50,WebFace\\(Softmax,CosFace,ArcFace)}
				&56.70\%&69.80\%&83.80\%\\ \hline\hline\hline
				
				\tabincell{c}{E-DI-M-FIM\\\cite{liu2016delving}\cite{xie2019improving}\cite{dong2018boosting}\cite{zhong2019adversarial}}&
				\tabincell{c}{ResNet-50,VGGFace2\\(Softmax,CosFace,ArcFace)}
				&67.19\%&76.99\%&86.49\%\\ \hline\hline
				
				\tabincell{c}{DFANet-\\E-DI-M-FIM}&
				\tabincell{c}{ResNet-50,VGGFace2\\(Softmax,CosFace,ArcFace)}
				&71.29\%&80.50\%&88.30\%\\ \hline\hline\hline
				
				\tabincell{c}{E-DI-M-FIM\\\cite{liu2016delving}\cite{xie2019improving}\cite{dong2018boosting}\cite{zhong2019adversarial}}&
				\tabincell{c}{ResNet-50,MS1M\\(Softmax,CosFace,ArcFace)}
				&68.50\%&76.29\%&83.59\%\\ \hline\hline
				
				\tabincell{c}{DFANet-\\E-DI-M-FIM}&
				\tabincell{c}{ResNet-50,MS1M\\(Softmax,CosFace,ArcFace)}
				&73.70\%&81.69\%&88.40\%\\ \hline
			\end{tabular}
		} 
	\end{center}
\end{table}

\subsection{Ablation Studies}
We conduct ablation experiments to study the impact of hyperparameters including the drop rate $p_d$ and the maximum number of iterations $N_{max}$, for a better understanding of the proposed DFANet. 

\begin{figure}[htbp]
	\center
	\includegraphics[width=0.9\linewidth]{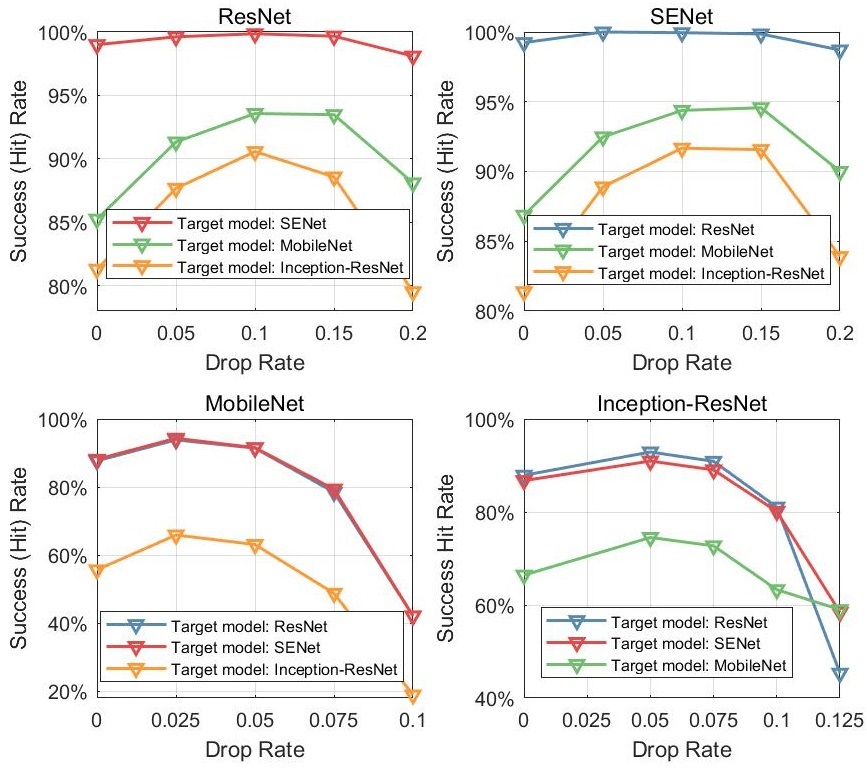}
	\caption{The success (hit) rates of the transferable adversarial attacks when varying the drop rate $p_{d}$. The adversarial examples are generated from model by DFANet-DI-M-FIM. Note that DFANet-DI-M-FIM degrades into DI-M-FIM when the drop rate is $p_{d}=0$. The source models are trained on CAISA-WebFace~\cite{Yi2014CASIA} supervised by the softmx loss, with different network architectures, namely, ResNet-50~\cite{he2016deep}, SENet-50~\cite{hu2018squeeze}, MobileNet~\cite{howard2017mobilenets}, and Inception-ResNet~\cite{szegedy2017inception}.}
	\label{fig:DCFNet_param_droprate}
\end{figure}

\textbf{Influence of the drop rate.} We first study the influence of the drop rate $p_{d}$ on the success (hit) rates. We generate transferable adversarial examples of the four models trained on CAISA-WebFace supervised by the softmx loss, with different network architectures, namely, ResNet-50, SENet-50, MobileNet and Inception-ResNet, using DFANet-DI-M-FIM. We set the maximum number of iterations $N_{max}$ to 1,500 and then change the drop rate $p_{d}$. Note that DFANet-DI-M-FIM degrades into the baseline method, DI-M-FIM, when the drop rate is $p_{d}=0$. 

The results are shown in Fig.~\ref{fig:DCFNet_param_droprate}. We find that the success (hit) rate first increases and then decreases with increasing the drop rate $p_{d}$. The optimal drop rate $p_{d}$ may vary across different network architectures. In addition, the change curves of similar network architectures ResNet-50 and SENet-50 are also similar. 

To figure out the reason behind Fig.~\ref{fig:DCFNet_param_droprate}, we should first consider how DFANet works. DFANet applies dropout layers to increase the diversity of surrogate models and obtain ensemble-like effects. The quality of the surrogate model in each iteration is decided by the drop rate $p_{d}$. If the drop rate is too small, the generated surrogate models in each iteration will become similar and the diversity of gradients will be insufficient, which will not provide significant improvement of the attack performance. However, if the drop rate is too high, the generated surrogate models will lose the basic recognition ability and the gradient of them will become pointless, which will decrease the attack performance. This finding indicates that a certain degree of randomness incorporated into the convolutional layers would help, but applying excessive randomness would have a negative effect.   
 
\begin{figure}[htbp]
	\center
	\includegraphics[width=1\linewidth]{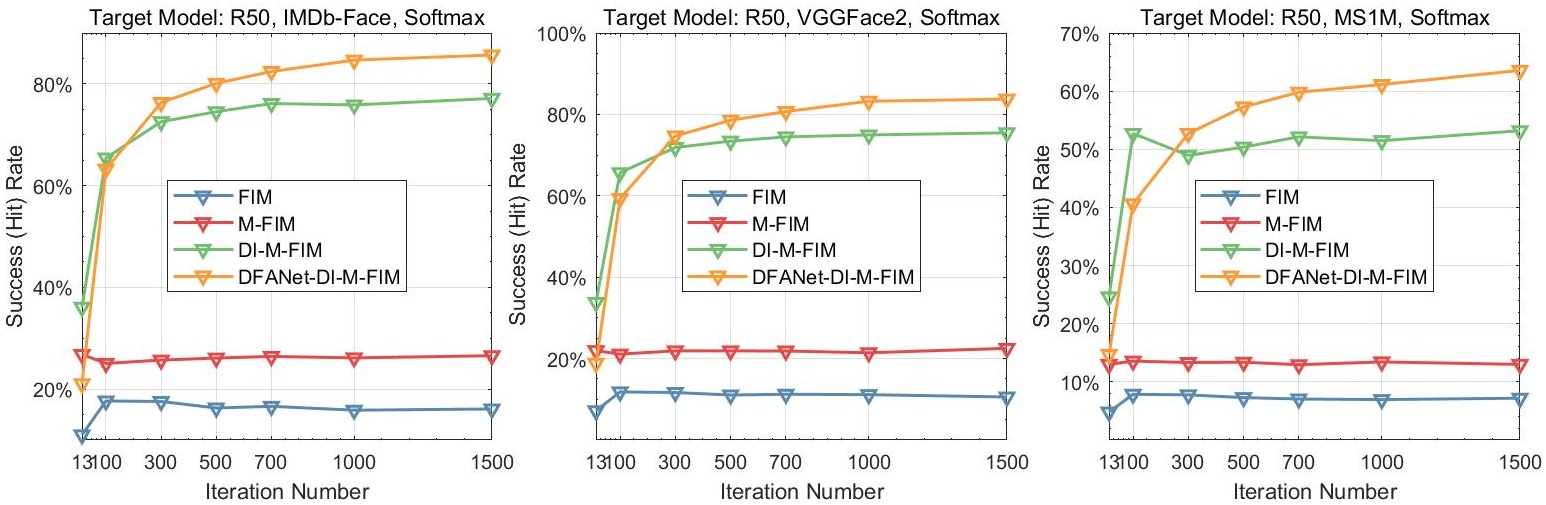}
	\caption{The success (hit) rates of transferable adversarial attacks when varying the maximum number of iterations $N_{max}$. The adversarial examples are generated by FIM, M-FIM, DI-M-FIM and DFANet-DI-M-FIM using the ResNet-50 model trained on CAISA-WebFace~\cite{Yi2014CASIA} and supervised by the softmax loss.}
	\label{fig:DCFNet_param_iteration}
\end{figure}

\textbf{Influence of the maximum number of iterations.} We then study the influence of the maximum number of iterations $N_{max}$ on the success (hit) rates. We generate transferable adversarial examples of ResNet-50 model trained on WebFace supervised by the softmx loss, using FIM, M-FIM, DI-M-FIM and DFANet-DI-M-FIM. Since we find that the optimal value of the drop rate $p_d$ for ResNet-50 is approximately 0.1, we maintain it as 0.1 and then change the maximum number of iterations $N_{max}$ from 13 to 1,500. 

The experimental results are shown in Fig.~\ref{fig:DCFNet_param_iteration}. The success (hit) rates for both DI-M-FIM and DFANet-DI-M-FIM improve as the maximum number of iterations $N_{max}$ increases. Although at the beginning, the success (hit) rate of DFANet-DI-M-FIM is lower, it gains more improvements and outperforms DI-M-FIM as $N_{max}$ increases. In addition, the trends in FIM and M-FIM are different from those in DI-M-FIM and DFANet-DI-M-FIM, where as the maximum number of iterations $N_{max}$ increases, the success (hit) rate remains almost unchanged. It is not hard to see the reason, with more iterations, the DFANet can further increase the diversity of the generated surrogate models and obtain better ensemble-like effects. 

\subsection{Discussion}
To achieve more intuitive understanding of transferable adversarial attacks and the proposed DFANet, we next interpret the intermediate generation process of adversarial example. 

\begin{figure}[htbp]
	\center
	\includegraphics[width=0.88\linewidth]{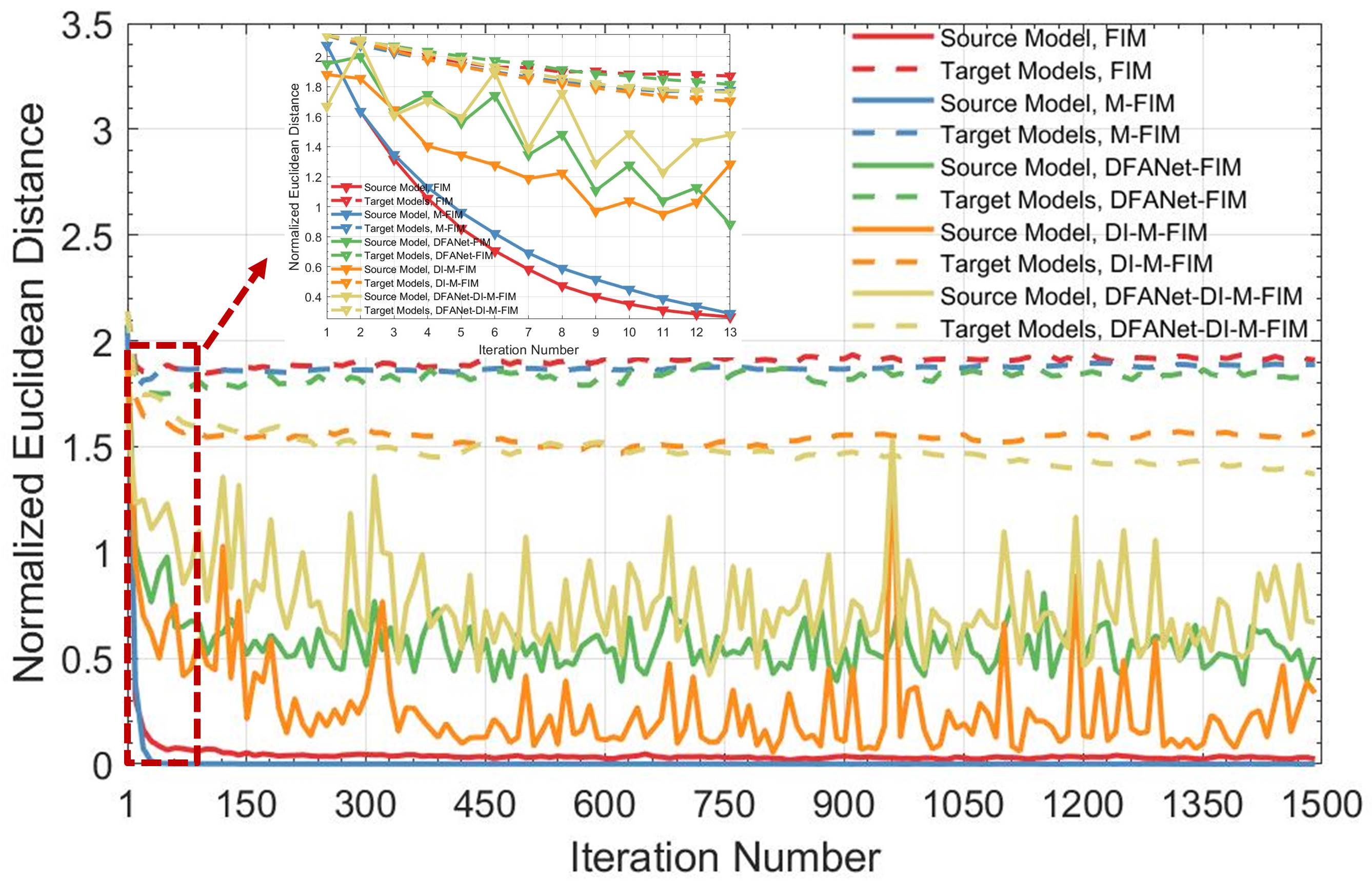}
	\caption{The average Euclidean distance of normalized deep features of 100 face pairs as the number of iterations increases. We provide two types of deep features: one type is extracted from the source model, while the other one is extracted from three target models. The adversarial examples are generated by FIM, M-FIM, DFANet-FIM, DI-M-FIM and DFANet-DI-M-FIM.}
	\label{fig:paper_dis}
\end{figure}

Since for deep face recognition, positive and negative pairs are judged according to the distances between their deep features, we observe the Euclidean distance of the normalized deep features in the generation process, which is also the objective loss function of feature-level attacks (referred to Equation~(\ref{fealoss1})(\ref{fealoss2})). First, we pick 100 face pairs randomly from the aforementioned 10,000 face pairs. Then, we generate adversarial examples from the ResNet-50 model trained on WebFace supervised by the softmax loss, using FIM, M-FIM, DFANet-FIM, DI-M-FIM and DFANet-DI-M-FIM. At the end of each iteration, we extract the deep features of the source model and three target models, which are ResNet-50 models trained on IMDb-Face, VGGFace2, MS1M, supervised by the softmax loss. For both source and target models, as the number of iterations increases, we record the average normalized Euclidean distances of the 100 face pairs in Fig.~\ref{fig:paper_dis}. 

We first focus on the average normalized Euclidean distances of the source model. For FIM and M-FIM, the average normalized Euclidean distances of the source models decrease constantly as the number of iterations increases. For DFANet-FIM, DI-M-FIM and DFANet-DI-M-FIM, although in the long term, the average normalized Euclidean distances of the source model decrease overall as the number of iterations increases; there are high fluctuations of these distances in a short term.

The fluctuations reflect the ensemble-like effects of the DFANet and DI method. Specifically, in all the iterations, if use the same model and the same input to calculate gradients, like gradients in FIM and M-FIM method, then the normalized Euclidean distances of the source model will decrease monotonically, because the adversarial perturbations are always updated towards the direction of decreasing the normalized Euclidean distances. DFANet obtain diversity of different surrogate models generated by dropout, while the DI method obtains diversity of the input images. These two type of variation will incorporate diversity to the gradients, therefore the normalized Euclidean distances of the source model will not decrease monotonically, but decrease with fluctuation. With the diversity of the gradients, these two type of variation will prevent overfitting to the single source model, and increase the possibility to adapt to more types of target models. 

Just because of this, we can find that, for DFANet-FIM, DI-M-FIM and DFANet-DI-M-FIM, the attack performance for target models is increased (normalized Euclidean distances are decreased); however, the attack performance for source model is decreased (normalized Euclidean distances are increased). If the adversarial examples are designed to adapt to more target models, then they cannot overfit to the one source model. It is important to incorporate diversity to gradients and DFANet can just be a method to realize this idea. In the figure, there exists a gap between the average normalized Euclidean distances for source models and those for target models under any method. We can see that for DFANet-DI-M-FIM, the performance of the source model is almost consistent with that of the target models, which reflects the advantage of the combination of DFANet and the strong baseline method, because the smaller these gaps are, the more transferable the generated adversarial examples are. 

\subsection{Comparison with Other Attack Methods}
In this section, we compare the proposed DFANet with other attack methods, including Skip Gradient Method (SGM)~\cite{wu2020skip} and GD-UAP~\cite{mopuri2018generalizable}. SGM is the up to date work to enhance the transferability of black box adversarial examples. SGM also makes subtle modifications to the surrogate models, but the surrogate model will not change in the generation process of adversarial examples. GD-UAP~\cite{mopuri2018generalizable} is an advanced universal adversarial attacks which can fool the deep models over multiple images, and across multiple tasks. 

To make fair comparison with SGM, we combine it with FIM, because we have demonstrated that feature level attacks are more effective than label level ones. We name the combination as S-FIM, and compare it with DFANet-FIM. The objective of GD-UAP launches attacks in feature level, which attempts to over-fire the neurons at multiple layers in order to deteriorate the extracted features. Therefore, we do not apply modifications to this method. 

Since GD-UAP can generate untargeted attacks, following the classic work on attacks towards face models~\cite{goswami2019detecting}, we conduct experiments on two public available face databases, namely, the Multiple Encounters Dataset (MEDS)~\cite{founds2011nist} and the Point and Shoot
Challenge (PaSC) database~\cite{beveridge2013challenge}. The MEDS-II database~\cite{founds2011nist} contains a total of 1,309 faces pertaining to 518 individuals. We use the metadata provided with the MEDS release 2 database to obtain a subset of 858 frontal face images from the database. Each of these images is matched to every other image and the resulting 855 $\times$ 855 score matrix is utilized to evaluate the verification performance. 

The PaSC database~\cite{beveridge2013challenge} contains still-to-still and video-to-video matching protocols. We use the frontal subset of the still-to-still protocol which contains 4,688 images pertaining to 293 individuals which are divided into equally sized target and query sets. Each image in the target set is matched to each image in the query set and the resulting 2344 $\times$ 2344 score matrix is used to determine the verification performance.

For evaluating performance of attack methods, we randomly select 50\% of the total images from each database and generate adversarial perturbations on them. For fair comparison between image-specific (FIM, S-FIM, DFANet-FIM) and universal adversarial attacks (GD-UAP), we remove the face pairs whose two images are both selected for generating adversarial examples since we find that universal attacks can lead two adversarial images become similar. In this way, all the face pairs have at most one adversarial examples. The new datasets with a half of adversarial examples are used to compute the new score matrices. The source model is the ResNet-50 model trained on CAISA-WebFace, supervised by ArcFace. Other nine models of different training databases, loss functions, and network architectures are used as target models.

\begin{figure}[htbp]
	\center
	\includegraphics[width=0.9\linewidth]{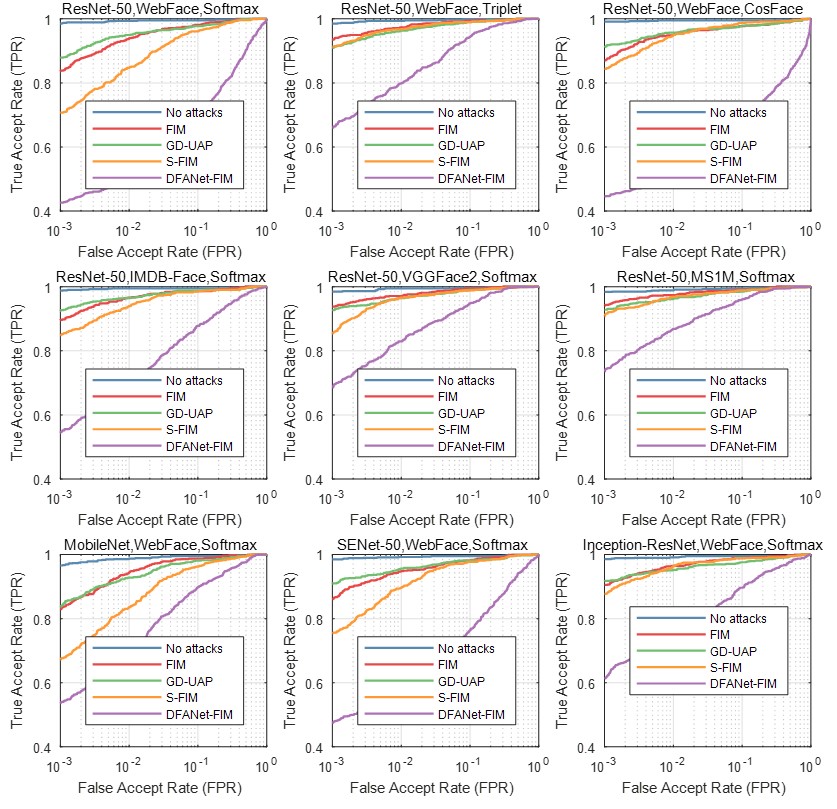}
	\caption{The ROC curves for nine target deep face models with different loss functions, training databases, network architectures, on the MEDS~\cite{founds2011nist} database respectively with no attacks, FIM, GD-UAP, S-FIM, and DFANet-FIM. For the attack method, lower is better.}
	\label{fig:MEDS_r50_webface_arc}
\end{figure}

\begin{figure}[htbp]
	\center
	\includegraphics[width=0.9\linewidth]{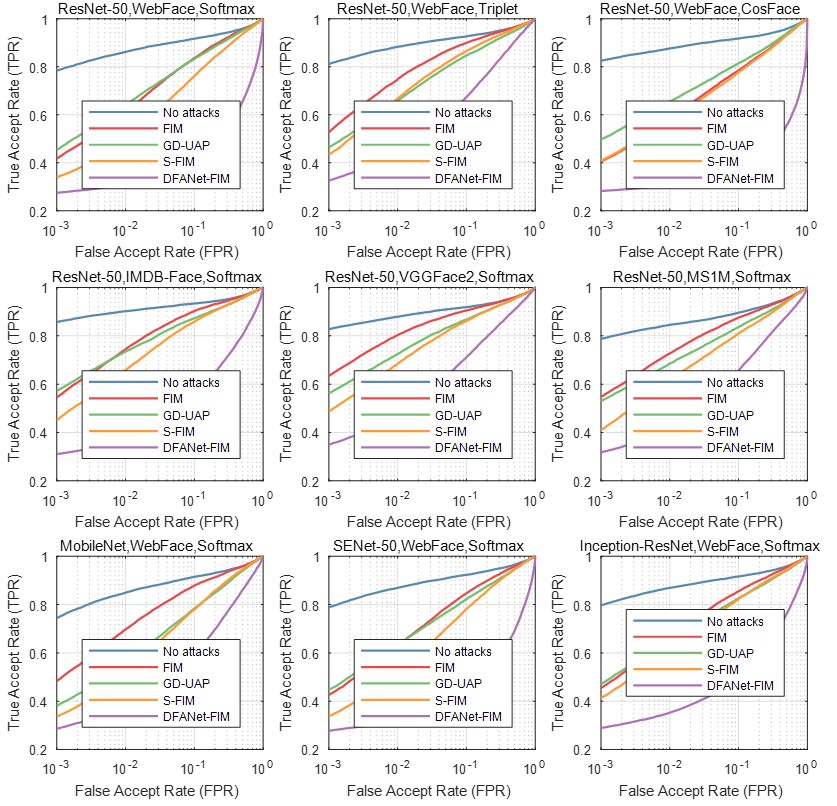}
	\caption{The ROC curves for nine target deep face models with different loss functions, training databases, network architectures, on the PaSC~\cite{beveridge2013challenge} database  respectively with no attacks, FIM, GD-UAP, S-FIM, and DFANet-FIM. For the attack method, lower is better.}
	\label{fig:PASC_r50_webface_arc}
\end{figure}

To obtain best performance of S-FIM, we experiment on MEDS database to select the important hyper-parameter for SGM, the decay parameter $\gamma$, which is related to the source model therefore we cannot use the empirical value as the original paper~\cite{wu2020skip}. We put the ablation study of the decay parameter in the Appendix. Finally, we choose the decay parameter $\gamma$ as 0.3 for the source model. GD-UAP is not sensitive for hyper-parameters, we optimize at the last layers of all residual blocks and the independent convolutional layers following the original paper~\cite{mopuri2018generalizable}. For better performance, we use the range prior calculated from MEDS database. We set the maximum number of iterations $N_{max}$ to 40,000 for GD-UAP, 1,500 for S-FIM and DFANet-FIM. For all attack methods, we generate untargeted attacks under maximum $L_{\infty}$ perturbation $\varepsilon=10$ with respect to pixel values in $\left[ 0,255 \right]$. 

The ROC curves on the MEDS database respectively with no attacks, FIM, GD-UAP, S-FIM, and DFANet-FIM are shown in Fig.~\ref{fig:MEDS_r50_webface_arc}, while the ROC curves on the PaSC database are shown in Fig.~\ref{fig:PASC_r50_webface_arc}. The lower are the ROC curves, better attack performance the black-box attacks have. We can observe that, DFANet-FIM has better attack performance than S-FIM and GD-UAP, which reflects DFANet-FIM can generate more transferable adversarial examples.

\subsection{TALFW Database}
We have thoroughly investigated adversarial attack methods against deep face recognition in this paper. With the help of transferability enhancement methods, the success (hit) rate can be as high as almost 90\%, which should raise security concerns for deep face models. Therefore, we aim to build a test benchmark to facilitate research on the robustness and generalization of face recognition. The Labeled Faces in the Wild (LFW)~\cite{LFWTech} database is a well-known test benchmark in the deep face recognition literature. Although the performance on the LFW database has been saturated, due to its ease of use and popularity, there may be a potential possibility for it to become an appropriate baseline for studying the robustness issue. In this section, based on the aforementioned transferable adversarial attack methods, we create the Transferable Adversarial LFW (TALFW)$\footnote{The website of TALFW database is \url{http://www.whdeng.cn/TALFW/index.html}, where the original images, protocols, codes and baselines are publicly available. You are encouraged to obtain the download links by sending emails.}$ database by adding noise imperceptible to human to the original LFW images. Since the only difference is the imperceptible noise, the evaluation protocol of TALFW is exactly the same as that of LFW, which will make it an easy-to-use and outstanding test database for the community. 

\begin{figure*}[htbp]
	\center
	\includegraphics[width=0.85\linewidth]{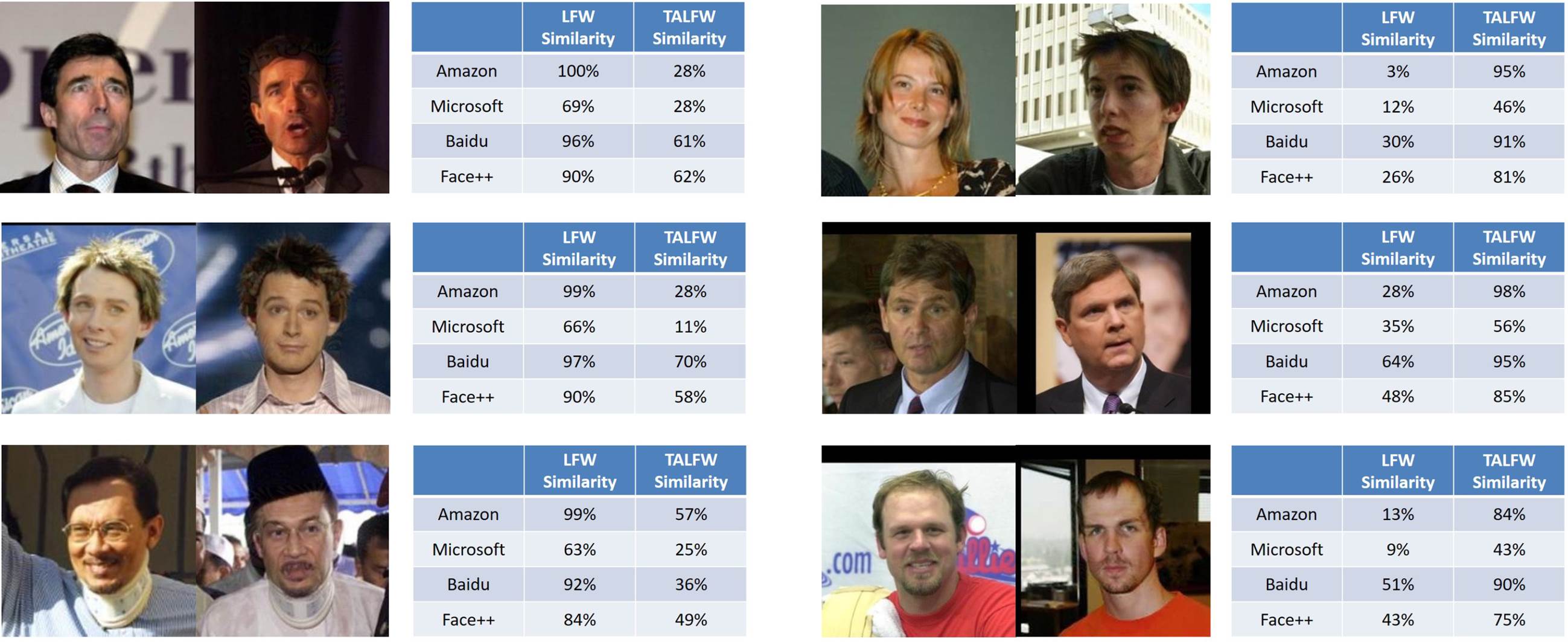}
	\caption{Three positive (left) and three negative (right) pairs in the TALFW database. The similarity scores of commercial APIs, namely, Amazon~\cite{Amazon}, Microsoft~\cite{azure}, Baidu~\cite{Baidu} and Face++~\cite{Face++} are also listed. The modification is nearly imperceptible to humans but can change the face similarity scores significantly in the black-box setting without queries.}
	\label{fig:subset}
\end{figure*}

Based on the aforementioned methods, we modify the original Labeled Faces in the Wild (LFW)~\cite{LFWTech} database using our private models. This modified database can be used to evaluate the robustness of mainstream open-sourced deep face models and commercial APIs. The original LFW database contains 13,233 face images of 5,749 identities. In the recent literature, the LFW database has been widely used to evaluate the performance of deep face models by testing on 3,000 positive and 3,000 negative face pairs, which involve 7,701 face images. Therefore, considering the evaluation protocol of the LFW database, the principal is to modify the face image in the pixel space slightly while significantly change the similarity of the corresponding face pairs in the deep feature space of unknown models. 

The steps to set up the Transferable Adversarial Labeled Faces in the Wild (TALFW) database are as follows. First, based on the greedy algorithm, we choose the minimum number of candidate face images to cover the maximum number of face pairs. Then we modify the candidate images in an imperceptible way.  Apart from the aforementioned transferable attack methods, we also use some techniques to reduce the visual impact of the modification. Since the size of face images in LFW database is 250$\times$250, we first generate adversarial perturbations on the aligned images using our private models. Then, we use the inverse transformation to transform the aligned adversarial examples to the original size, where the background information is from the original image in LFW. In this way, the generated adversarial images can be applied to a variety of deep face models with different aligned principles. An examples is shown in Fig.~\ref{fig:intro_pair}, where the generated adversarial perturbation has black paddings. 

In total, 4,069 face images are modified and compared with the original LFW database which has 3,000 positive and 3,000 negative face pairs. Additionally, we evaluate the robustness of the four commercial APIs and four state-of-the-art (SOTA) open-sourced models on the TALFW database and the original LFW database. Fig.~\ref{fig:subset} shows three positive (left) and three negative (right) face pairs in the TALFW database. We also list the similarity score of the four commercial APIs: Amazon~\cite{Amazon}, Microsoft~\cite{azure}, Baidu~\cite{Baidu} and Face++~\cite{Face++}. From the figures, the modification is nearly imperceptible to humans but can change the face similarity score significantly in the black-box setting without queries. More face pairs in TALFW database can be referred to the Appendix. 

We first test open source models of SOTA algorithms, \emph{i.e}., the center-loss~\cite{wen2016discriminative}, SphereFace~\cite{liu2017sphereface}, VGGFace2~\cite{cao2018vggface2} and ArcFace~\cite{deng2019arcface}. The ArcFace (ResNet-100) model has reported SOTA performance on several benchmarks including YTF, MegaFace challenge, and IJB-C~\cite{deng2019arcface}. We use MTCNN for face detection and strictly follow the preprocessing steps of the original algorithms. Compared with the original images in the LFW database, the perturbed images in the TALFW database have no influence on the accuracy and reliability of face detection. The accuracy of open-sourced models of the four SOTA algorithms on the LFW and TALFW databses is listed in the first cell of Table~\ref{table:result}. From the experimental results, there indeed exists a striking gap between the accuracy on the LFW and TALFW databases, which reflects that even the SOTA algorithms for deep face recognition are extremely vulnerable to transferable attacks.  

Then we tested LFW and TALFW on the commercial APIs including Amazon~\cite{Amazon}, Microsoft~\cite{azure}, Baidu~\cite{Baidu} and Face++~\cite{Face++}. Specifically, since the TALFW database is generated based on transferability, we obtain the similarity score only by once calling without any query feedback. We also have no knowledge about the whole pipelines of the commercial APIs. We only need to give the original images in both the LFW and TALFW database to the commercial APIs directly without any image preprocessing, and then we get the similarity score. The performance of the four commercial APIs on the LFW and TALFW databases is listed in the second cell of Table~\ref{table:result}. All the commercial APIs deteriorate seriously when transferring from the LFW database to the TALFW database, which reflects the idea that transferable adversarial attacks seriously threaten commercial face APIs.

\begin{table}
	\renewcommand\arraystretch{1.0}
	\newcommand{\tabincell}[2]{\begin{tabular}{@{}#1@{}}#2\end{tabular}}
	\caption{Evaluation Results of the Commercial APIs, SOTA algorithms and efensive models.}
	\label{table:result}
	\begin{center}
		\scalebox{1}{
			\begin{tabular}{|c|c|c|c|}
				\hline
				&Model&LFW&TALFW \\ \hline\hline
				\multirow{5}{*}{\tabincell{c}{SOTA\\Algorithms}}
				& Center-loss~\cite{wen2016discriminative}& 98.78 & 70.65\\ \cline{2-4}
				& SphereFace~\cite{liu2017sphereface}& 99.27 & 62.47\\ \cline{2-4}
				& VGGFace2~\cite{cao2018vggface2}& 99.43 & 71.47\\ \cline{2-4}
				& ArcFace (MobileNet)~\cite{deng2019arcface}& 99.35 & 50.77\\ \cline{2-4}
				& ArcFace (ResNet-100)~\cite{deng2019arcface}& 99.82 & 63.45\\
				\hline\hline
				\multirow{5}{*}{\tabincell{c}{Commercial\\APIs}}
				& Amazon~\cite{Amazon}& 99.47 & 69.28 \\ \cline{2-4} 
				& Microsoft~\cite{azure} &98.12& 70.93 \\ \cline{2-4}
				& Baidu~\cite{Baidu}& 97.72 & 72.07 \\ \cline{2-4}
				& Face++~\cite{Face++}& 96.95 & 73.90 \\ \cline{2-4}
				& Fusion of four APIs& 99.65 & 72.33 \\ \hline\hline
				
				
				\multirow{5}{*}{\tabincell{c}{Defensive\\Methods}}
				& No Defense &99.78&54.15\\ \cline{2-4}
				& JPEG Encoding~\cite{Kurakin17}&99.55&73.93\\ \cline{2-4}
				& Gaussian Blur~\cite{Kurakin17}&99.57&77.95\\ \cline{2-4}
				& Selective Dropout~\cite{goswami2019detecting}&99.52&72.77\\ \cline{2-4}
				& Adversarial Training~\cite{Kurakin17atscale}& 99.62& 82.17\\ 
				\hline
			\end{tabular}
		}
	\end{center}
\end{table}

\begin{figure}[htbp]
	\center
	\includegraphics[width=0.75\linewidth]{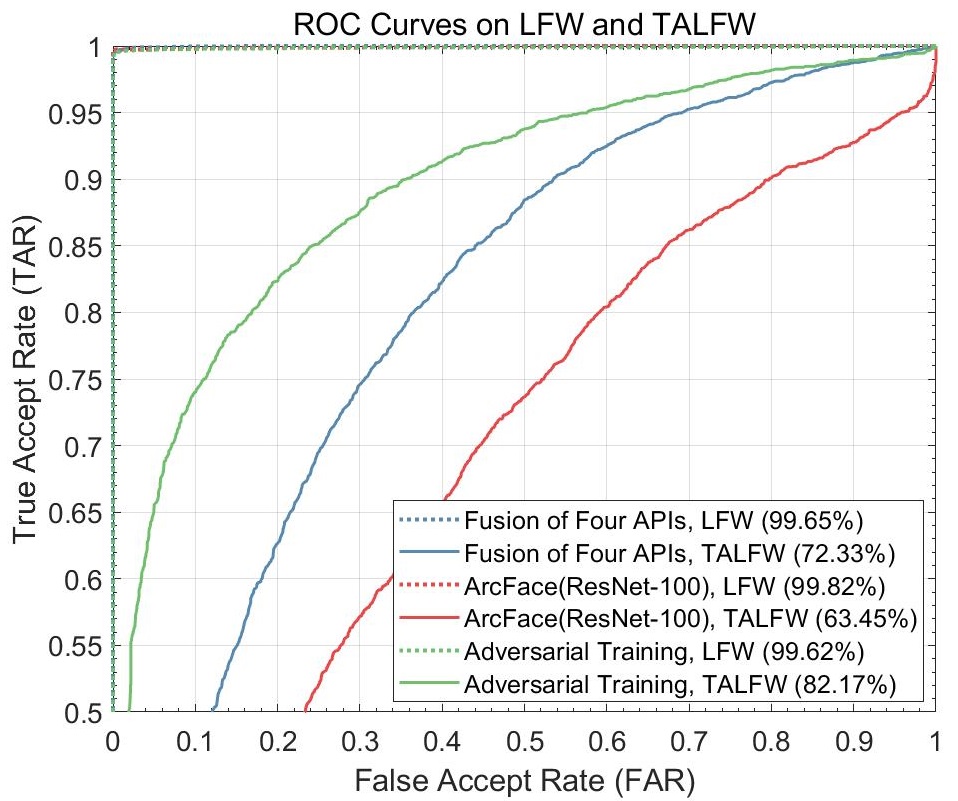}
	\caption{Comparison of the LFW and TALFW databases. We select some algorithms as example here: the fusion of commercial APIs (Amazon~\cite{Amazon}, Microsoft~\cite{azure}, Baidu~\cite{Baidu} and Face++~\cite{Face++}), the SOTA model trained on MS1M with ResNet-100 supervised by ArcFace, and a defensive method by adversarial training~\cite{Kurakin17atscale}. }
	\label{fig:roc_lfw_talfw}
\end{figure}

Furthermore, we test some defensive methods including JPEG encoding~\cite{Kurakin17}, Gaussian blur~\cite{Kurakin17}, Selective Dropout~\cite{goswami2019detecting} and adversarial training~\cite{Kurakin17atscale}. The compared no-defense model is a ResNet-50~\cite{he2016deep} model trained on MS-Celeb-1M~\cite{Guo16MS} with ArcFace loss~\cite{deng2019arcface}. There has been a consensus that the improvement in robustness would bring performance degradation on clean test images~\cite{Kurakin17,tsipras2018robustness}. Since we aim to evaluate the robustness of deep face models while at the same time keeping the recognition performance of original images at a relatively high level, we check the performance of defensive models on the TALFW database while recommend keeping the accuracy on the LFW database no less than 99\%. We determin the hyperparameters for prospective defensive methods and choose models for adversarial training methods following this rule. For JPEG encoding, the JPEG quality is chosen to be level 20 (out of 100); and for the Gaussian blur, the kernel size is set to 5 with standard deviation 2. For Selective Dropout, we make an ablation study on the hyper-parameters, the top $\gamma$ layers and the top $\kappa$ fraction of affected filters, which is listed in the Appendix. Finally, we choose $\gamma=4$ and $\kappa=0.05$. Adversarial examples incorporated into adversarial training are generated by FIM since they are more effective in face models than label-level methods. From the results, we find that compared with the original model, which only has 54.15\% accuracy on the TALFW database, although defensive methods can improve the performance to different degrees, the performance gap between the LFW and TALFW databases still exists, which reflects the idea that transferable adversarial attacks can be alleviated but there is still plenty of scope to push the corresponding techniques further.

We have evaluated the robustness of mainstream open-sourced deep face models, commercial APIs and defensive methods on the LFW and TALFW databases. Some typical comparison ROC curves are selected in Fig.~\ref{fig:roc_lfw_talfw}. Overall, the severe performance degradation from the LFW to TALFW databases clearly shows the vulnerability of deep face models.

\section{Conclusion}
As the recognition performance of deep face models improves, robustness and generalization have become increasingly essential and crucial. In this paper, we study applicable transferable adversarial attacks against deep face recognition. We first find a baseline method by exploring the attack methods from the label-level to the feature-level and demonstrate empirically that iterative feature-level attacks are more effective and transferable. We find that it is difficult to guarantee successful attacks against deep face models with basic iterative adversarial attacks. Therefore, we study transferability enhancement methods and propose DFANet to further improve transferability by increasing the possible settings of the model parameters and obtain ensemble-like effects of generated surrogate networks in the iterative steps. The experimental results demonstrate that the proposed method can be combined with various networks and existing transferability enhancement methods, which achieve effective black-box adversarial attacks against deep face models with different training databases, loss functions and network architectures. Based on the aforementioned study, we contribute a test database, TALFW, to help study the robustness and defensibility of deep face models. The hope is that TALFW database could raise much attention from researchers and industries in a timely manner.

\section*{Acknowledgment}
This work was supported by National Key R\&D Program of China (2019YFB1406504), and supported by BUPT Excellent Ph.D. Students Foundation CX2020201.

\ifCLASSOPTIONcaptionsoff
  \newpage
\fi

\bibliographystyle{IEEEtran}
\bibliography{dfanet}

\end{document}